\documentclass[letterpaper]{article} 
\usepackage[draft]{aaai25}  
\usepackage{times}  
\usepackage{helvet}  
\usepackage{courier}  
\usepackage[hyphens]{url}  
\usepackage{graphicx} 
\usepackage{natbib}  
\usepackage{caption} 
\usepackage{algorithm}
\usepackage{algorithmic}

\usepackage{amsmath}
\usepackage{amsfonts}
\usepackage{booktabs}
\usepackage{multirow}
\usepackage{enumitem}
\usepackage{bm}
\usepackage[most]{tcolorbox}
\tcbuselibrary{breakable}

\usepackage{newfloat}
\usepackage{listings}
\DeclareCaptionStyle{ruled}{labelfont=normalfont,labelsep=colon,strut=off} 
\floatstyle{ruled}
\newfloat{listing}{tb}{lst}{}
\floatname{listing}{Listing}
\title{IntelliCare: Improving Healthcare Analysis with Variance-Controlled \\ Patient-Level Knowledge from Large Language Models}
\author{
    Zhihao Yu$^1$, Yujie Jin$^1$, Yongxin Xu$^1$, Xu Chu$^{1,2,3}$, Yasha Wang$^{1,2}$, Junfeng Zhao$^{1,2}$
}
\affiliations{
    $^1$School of Computer Science, Peking University \\
  $^2$National Research and Engineering Center of Software Engineering, Peking University \\
  $^3$Center on Frontiers of Computing Studies, Peking University, Beijing, China\\
  yuzhihao@stu.pku.edu.cn, chu\_xu@pku.edu.cn, wangyasha@pku.edu.cn \\

}

\begin{document}

\maketitle

\begin{abstract}
    While pioneering deep learning methods have made great strides in analyzing electronic health record (EHR) data, they often struggle to fully capture the semantics of diverse medical codes from limited data. The integration of external knowledge from Large Language Models (LLMs) presents a promising avenue for improving healthcare predictions. However, LLM analyses may exhibit significant variance due to ambiguity  problems and inconsistency issues, hindering their effective utilization. To address these challenges, we propose IntelliCare, a novel framework that leverages LLMs to provide high-quality patient-level external knowledge and enhance existing EHR models. 
    Concretely, IntelliCare identifies patient cohorts and employs task-relevant statistical information to augment LLM understanding and generation, effectively mitigating the ambiguity problem. Additionally, it refines LLM-derived knowledge through a hybrid approach, generating multiple analyses and calibrating them using both the EHR model and perplexity measures.
    Experimental evaluations on three clinical prediction tasks across two large-scale EHR datasets demonstrate that IntelliCare delivers significant performance improvements to existing methods, highlighting its potential in advancing personalized healthcare predictions and decision support systems. Our code is available at \url{https://github.com/yzhHoward/IntelliCare}.
\end{abstract}

%

\begin{figure}[h]
    \centering
    \includegraphics[width=1\columnwidth]{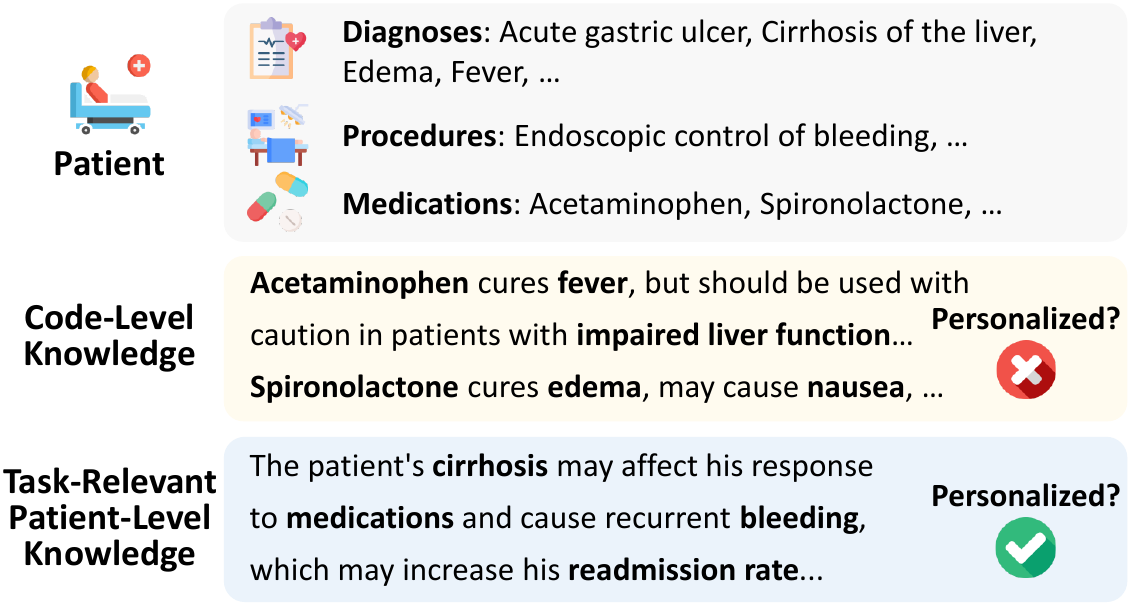}
    \caption{
        By combining with patient information, LLMs can provide the most relevant knowledge to the patient, rather than listing all information related to a certain medical code, although it might be irrelevant. Thus, employing analysis of patients as external knowledge with LLM avoids the overhead of filtering knowledge. Notably, LLM can provide unique insights in conjunction with clinical tasks and deliver more accurate external knowledge. 
    }
    \label{intro}
    \vspace{-0.25cm}
\end{figure}

\section{Introduction}
The widespread adoption of electronic health record (EHR) systems has led to an exponential growth in medical data, particularly in intensive care units (ICUs). These EHRs, containing crucial information such as patient conditions, surgical procedures, and medication regimens, have become invaluable resources for describing patient health status and predicting outcomes. This wealth of data has attracted significant attention from both computer scientists and medical researchers, spurring the development of various deep learning methods to analyze EHR data \cite{choi2016retain,baytas2017patient,bai2018interpretable,song2018attend,choi2018mime,ashfaq2019readmission,luo2020hitanet}.
However, EHR data typically comprises numerous medical codes, such as International Classification of Diseases (ICD) and National Drug Code (NDC), which present a challenge due to their vast variety. Consequently, these methods may struggle to fully grasp the semantics of each code from limited data. To address this issue, recent studies \cite{choi2020learning,lu2021collaborative,ye2021medpath,ye2021medretriever,lu2022context,xu2023vecocare,yang2023kerprint} have attempted to incorporate external knowledge, such as knowledge graphs, to compensate for the model's knowledge gaps.

Large Language Models (LLMs), which are trained on vast corpora of literature, possess powerful analytical and reasoning capabilities that could be harnessed as external knowledge for EHR analysis. Previous works \cite{jiang2024graphcare,xu2024ram} have shown the potential of LLMs to enhance existing models for providing knowledge for EHR data with medical codes. However, these approaches are limited to providing knowledge at the medical code level and fail to offer patient-level insights. In contrast, LLMs can provide a comprehensive understanding of patient health conditions to generate personalized analyses \cite{cui2024llms,zhu2024prompting}.
In Figure \ref{intro}, we provide a real-world case to illustrate the potential benefits in employing LLMs to analyze patient health conditions.
To fully leverage the capabilities of LLMs, in this work, we propose a general framework that prompts LLMs to analyze individual patient health conditions as a source of external knowledge, serving to enhance existing healthcare models and improve their prediction quality. While this approach seems intuitive, it is important to note that the analyses from LLMs can exhibit significant variance, primarily due to two key factors:

\begin{itemize}[leftmargin=*]
    \item \textbf{LLM analyses can be ambiguous in clinical tasks:} Previous attempts to use LLMs for predicting patient health states have shown limited performance due to the complexity of individual patient conditions \cite{cui2024llms,wang2024augmented}. Although augmenting LLMs with knowledge graphs and other data sources can mitigate errors in medical knowledge and improve performance, \cite{xu2024ram,zhu2024realm}, LLMs may still lack a comprehensive understanding of healthcare prediction tasks and provide vague responses, such as \textit{"the patient may stay longer in the hospital"}. This deficiency can lead to potentially misleading and unreliable information, resulting in variance in the generated analyses. 
    \item \textbf{Inconsistency in LLM analyses:} Research has indicated that most existing LLMs may exhibit high variability and inconsistency \cite{dong2024statistical,zhao2023knowing,xu2024knowledge}. For the same patient, multiple LLM queries may yield vastly different assessments, such as describing the patient's condition as complex (neutral) or critical (negative). This inconsistency poses significant challenges to patient prognosis, introducing substantial variance during integrating with healthcare models, compromises model stability, and affects predictive accuracy.
\end{itemize}

By jointly considering these issues, we introduce IntelliCare, a novel framework designed to integrate LLMs' intelligence into healthcare prediction by facilitating high-quality patient-level external knowledge provision from LLMs while reducing the variance in analyses. 
IntelliCare automatically identifies potential patient cohorts and utilizes task-relevant information within these cohorts to help LLMs better understand potential patient conditions. In this way, LLMs can compare the patient with similar cohorts and generate more reliable knowledge, which effectively addresses the ambiguity problem in LLM analyses.
Moreover, we implement an innovative LLM knowledge refinement method: We generate numerous analyses by querying the LLM multiple times, collectively serving as external knowledge. We then refine this LLM-derived knowledge from two perspectives, using both the EHR analysis model and perplexity measures, which help alleviate inconsistency issues and reduce variance in the external knowledge.

Our contributions can be summarized as follows:

\begin{itemize}[leftmargin=*]
    \item We propose a general framework that enhances the performance of existing models by leveraging patient-level external knowledge to augment EHR analysis.
    \item We design patient cohort identification and hybrid analysis refinement to control the variance in LLM knowledge, addressing the ambiguity and inconsistency issues.
    \item We validate the effectiveness of our proposed model on three tasks across two real-world EHR datasets (MIMIC-III and MIMIC-IV). Through extensive experimentation, we demonstrate that all the components contribute to performance improvements. Our results indicate that IntelliCare consistently outperforms existing methods across multiple evaluation metrics.
\end{itemize}

\section{Related Works}
\textbf{Healthcare Predictive Models without External Knowledge.} As EHR data has become increasingly recognized as a valuable resource in the clinical domain, various deep learning techniques have shown excellent performance in healthcare predictions by exploiting its structured nature \cite{choi2016multi,choi2020learning,ye2020lsan,ma2020adacare,gao2020stagenet,ma2021distilling,cai2022hypergraph} based on recurrent neural networks, Transformer architectures \cite{vaswani2017attention}, and graph neural networks (GNNs). Recent advances in healthcare predictive models have focused on capturing feature correlations \cite{luo2020hitanet,ma2022patient} via sophisticated network architectures, and improving interpretability \cite{ma2020concare,xu2022counterfactual,ma2023mortality}. For example, \cite{ma2022patient} compact representation by imposing a correlational sparsity prior to the correlations of medical feature pairs. \cite{xu2022counterfactual} propose a counterfactual learning model to capture higher-order interactions among medical codes while providing interpretations. However, due to the complexity of EHR data, these models may struggle to fully understand the semantics of each code, leading to suboptimal performance.

\textbf{External Knowledge Integration in Healthcare Prediction.} To address the above limitation, researchers have attempted to incorporate external knowledge, such as medical ontologies or knowledge graphs, to improve representation learning, thereby enhancing prediction performance. For example, \cite{ye2021medretriever} retrieve unstructured clinical text for each medical code to introduce reliable guides. \cite{choi2017gram,ma2018kame,yin2019domain,lu2021collaborative,tan2022metacare} fuse the ancestor node representations of medical ontologies through an attention mechanism. \cite{ye2021medpath,gao2022medml,xu2023seqcare,yang2023kerprint,xu2023vecocare} utilize knowledge graphs to introduce the relationships between medical codes. 
Thanks to the development of LLMs, \cite{jiang2024graphcare} employ them to build triples between medical entities to enhance existing knowledge graphs.
Nevertheless, this approach limits the potential of LLMs by only allowing them to provide information for medical codes and fails to provide patient-level knowledge. In the perspective of leveraging LLMs, \cite{cui2024llms,wang2024augmented,zhu2024prompting} have shown that LLMs can be used to assess patients' health conditions, but their performance is not satisfactory due to lack of understanding of clinical tasks. \cite{shoham2023cpllm} attempt to finetune LLMs but show sub-optimal performance compared to the latest EHR models. Recently, \cite{xu2024ram,zhu2024realm} adopt LLMs to summarize external medical databases, including knowledge graphs, literature, and encyclopedias, to provide more comprehensive knowledge for each medical entity. However, these two methods do not consider the variance of LLM knowledge which may lead to misunderstanding information and cannot provide personalized knowledge as well. In contrast, IntelliCare conducts multiple analyses for each patient using LLMs as external knowledge and calibrates the knowledge to mitigate the variance and improve healthcare prediction.

\section{Methodology}
\begin{figure*}[h]
    \centering
    \includegraphics[width=1.99\columnwidth]{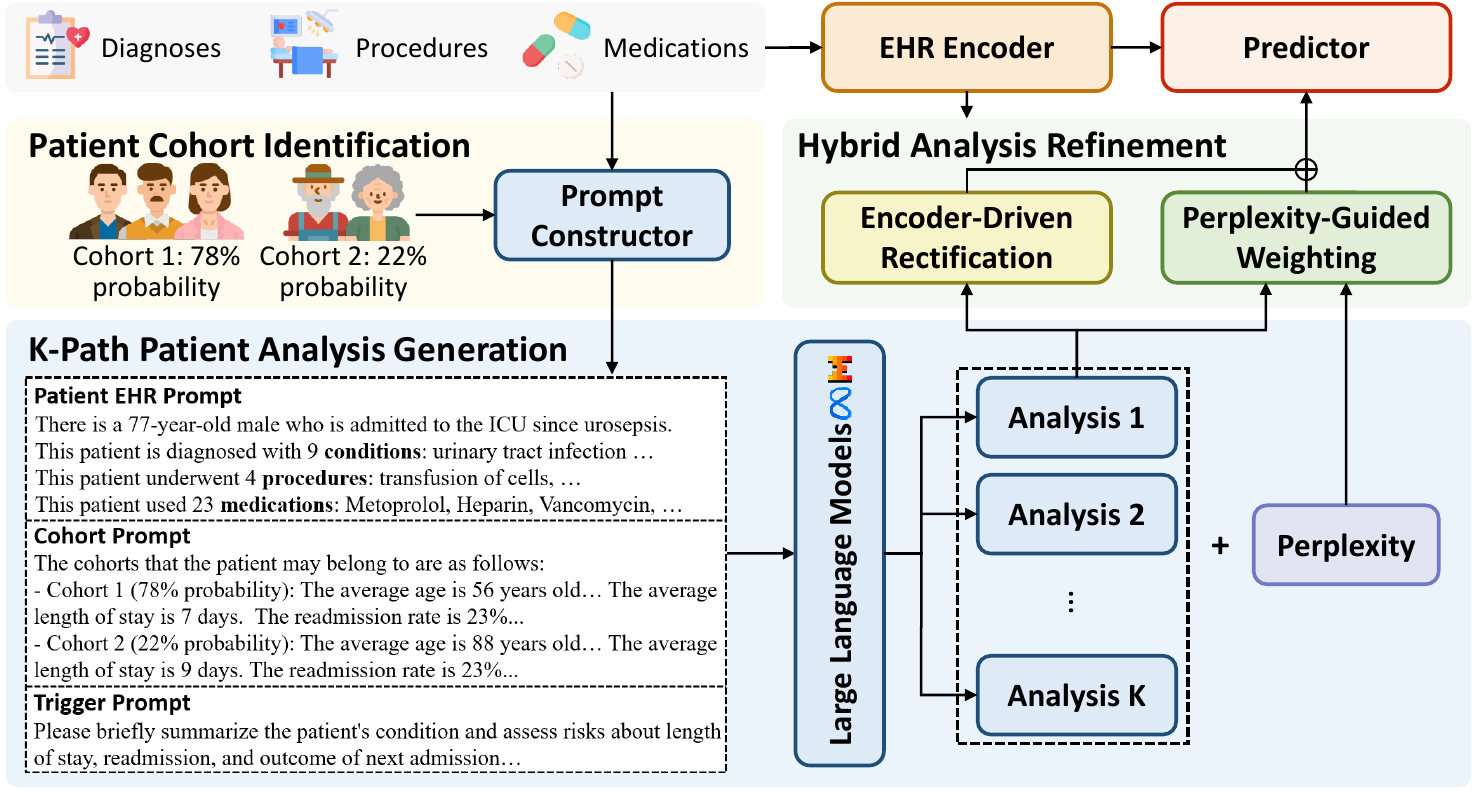}
    \caption{\textbf{Overview of IntelliCare}. Given the electronic medical records of a patient, IntelliCare constructs prompt via the records and task-relevant information within the cohorts to which the patient may belong. We then generate multiple analyses from LLMs for the patient and design a hybrid analysis refinement to calibrate this knowledge. Finally, we combine the refined knowledge with the existing trained EHR model to improve their prediction performance.}
    \label{framework}
\end{figure*}

In this section, we introduce IntelliCare, a novel framework that integrates LLMs to provide intelligent patient-level knowledge for enhancing existing healthcare predictions. Figure \ref{framework} provides an overview of the information flow and highlights our key design elements.

\subsubsection{Problem Formulation} The medical records of a patient are composed of demographic information consisting of age, gender, etc., and sets of conditions, procedures, and medications encoded by medical codes, respectively. The sizes of the sets vary across patients. The goal of healthcare prediction is to forecast the future health status of the patient, such as readmission, mortality, or length of stay in the ICU.

\subsection{Patient Cohort Identification}
Considering that LLMs may lack a comprehensive understanding of healthcare prediction tasks, we propose identifying potential patient cohorts and utilizing task-relevant information within these cohorts to help LLMs better understand patient conditions. To prevent potential information leakage, we perform unsupervised clustering exclusively on the training set. While this approach may seem straightforward, several challenges arise in identifying patient cohorts: 
\begin{itemize}[leftmargin=*]
    \item \textbf{High dimensionality:} The diverse and sparse vector of medical codes poses significant challenges for traditional clustering methods, which are computationally expensive and struggle to accurately assess similarities in high-dimensional one-hot spaces.
    \item \textbf{Unknown number of cohorts:} Popular unsupervised clustering methods, such as K-Means and Gaussian Mixture Model (GMM), require specifying the overall number of clusters. However, there is insufficient prior knowledge to determine this number accurately.
    \item \textbf{Multi-cohort assignment:} Patients in ICUs often have complex conditions and may suffer from multiple diseases simultaneously. Consequently, a single patient may belong to multiple cohorts. Most clustering methods assign each patient to a single cluster, which may not accurately reflect the patient's health condition.
\end{itemize}

To address the high-dimensionality challenge, we employ Uniform Manifold Approximation and Projection (UMAP) \cite{mcinnes2018umap}, a dimensionality reduction technique based on manifold learning. During the dimensionality reduction process, we use the Dice-Sorensen distance \cite{dice1945measures,sorensen1948method} to measure the similarity between patients, as it effectively quantifies the similarity between two sets. To tackle the latter two challenges, we utilize a Variational Bayesian Gaussian Mixture Model (VBGMM) \cite{corduneanu2001variational, bishop2006pattern}. Given a dataset $\mathcal{D}$, the VBGMM involves the use of a Gaussian mixture model having a fixed number of potential components (corresponding to the maximum possible number of components), in which the mixing coefficients $\pi$ are optimized to maximize the marginal data likelihood $p(\mathcal{D}|\pi)$. Due to the intractable marginalization, the VBGMM uses a variational approximation scheme based on the maximization of a lower bound on $\ln{p(\mathcal{D}|\pi)}$. This optimization procedure causes the mixing coefficients corresponding to unwanted components to approach zero, thereby automatically determining the optimal number of components. This method allows for a flexible number of cohorts and naturally handles multi-cohort assignments by providing probabilistic memberships for each patient across multiple clusters.

For each cohort, we extract cohort-specific statistical information related to the patient (including average age, gender, and number of conditions, procedures, and medications) and prediction tasks, such as average length of stay, readmission rate, and mortality rate. For a given patient, we calculate their probabilistic memberships across all cohorts. We then select the cohorts with probabilities above a threshold $\theta$ for the patient and use the corresponding information to augment the LLM's understanding of the patient's condition, which provides a broader context of similar patient groups, potentially improving their ability to analyze individual patient cases and generate more reliable and task-relevant knowledge.

To leverage the analytical capabilities of LLMs for patient-specific EHR analysis, we construct prompts comprising three essential components:
\begin{itemize}[leftmargin=*]
    \item \textbf{Patient EHR Prompt:} This includes demographic information, conditions, procedures, and medications. Medical codes are converted to their corresponding textual descriptions to enhance readability and comprehension for LLM.
    \item \textbf{Cohort Prompt:} We include information about the cohorts to which the patient likely belongs. This cohort-based information provides the LLM with a broader context about similar patient groups, potentially improving its ability to analyze the patient case and generate more reliable and task-relevant knowledge.
    \item \textbf{Trigger Prompt:} We provide specific instructions for the LLM to generate a summary and analysis of the patient's health condition related to clinical tasks.
\end{itemize}
When generating analyses, we sample $K$ responses from the LLM. Simultaneously, we calculate the perplexity using the token probabilities during generation. This information will be used to refine the LLM's knowledge in subsequent steps.

\subsection{Hybrid Analysis Refinement}
To address the instability of LLM's knowledge, we propose a hybrid analysis calibration method that utilizes information from both the existing EHR encoder and perplexity measures to independently calibrate the knowledge from two different perspectives.

\subsubsection{Encoder-Driven Rectification}

For the EHR encoder-driven calibration, we assume that higher consistency
between the embedding from the EHR encoder and the analysis indicates more accurate knowledge. Thus, we compute the correlation between them and adaptively select external knowledge that aligns well with the EHR encoder output.

Specifically, denoting $\bm{h}$ as the embedding from the EHR encoder, and $\mathcal{Z}=\{\bm{z}_1, \bm{z}_2, ..., \bm{z}_K\}$ as the embeddings of the LLM analyses extracted by the pre-trained language model such as BERT-like models, we calculate the correlation coefficient between $\bm{h}$ and each analysis $\bm{z}_k$. We then use these weights to compute a weighted sum of $\bm{z}_k$, resulting in more stable and accurate knowledge. To determine the weights, we use the $\mathtt{sigmoid}$ function instead of the commonly chosen $\mathtt{softmax}$. $\mathtt{sigmoid}$ assigns weight independently to each analysis, allowing for individual evaluation and preventing the dilution of different analyses. Specifically, it flexibly assigns low weights to all analyses if none are helpful, whereas softmax merely normalizes them.

Let $\bm{Q} \in \mathbb{R}^{d_h \times d_z}, \bm{K} \in \mathbb{R}^{d_z \times d_z}, \bm{V} \in \mathbb{R}^{d_z \times d_z}$ be three parameter matrices, where $d_h$ and $d_z$ are the output dimensions of the EHR encoder and LLM, respectively. The attention weight $\alpha$ is calculated as follows:
\begin{equation}
\alpha = \mathtt{sigmoid}(\frac{\bm{Q} \bm{h} \cdot \bm{K} \bm{z}_k^T}{\sqrt{d_z}})
\end{equation}
The encoder-driven rectified knowledge is obtained by:

\begin{equation}
    \bm{z}_e = \sum_{k=1}^{K} \alpha_k \cdot \bm{V} \bm{z}_k
\end{equation}

\subsubsection{Perplexity-Guided Weighting}

For the perplexity-driven calibration, we leverage the perplexity \cite{jelinek1977perplexity} of the LLM during generation to calibrate its knowledge. Considering that higher perplexity indicates greater uncertainty in the LLM's generated content, we calculate the perplexity weight as:
\begin{equation}
    \beta = \frac{1}{\mathtt{log}(p)},
\end{equation}
where $p$ is the perplexity of the generated analyses.

We then compute a weighted sum of the analyses $\bm{z}_k$ using the perplexity weights $\beta$, resulting in the perplexity-driven weighted knowledge:
\begin{equation}
    \bm{z}_p = \sum_{k=1}^{K} \bm{W} \beta_k \cdot \bm{z}_k,
\end{equation}
where $\bm{W} \in \mathbb{R}^{1 \times 1}$ is a transformation parameter matrix.

\subsubsection{Knowledge Fusion}

Finally, we fuse the encoder-driven calibrated knowledge $\bm{z}_e$ and the perplexity-guided weighting knowledge $\bm{z}_p$ to obtain the final calibrated knowledge by concatenating them and applying a linear transformation with layer normalization \cite{ba2016layer}:

\begin{equation}
    \bm{z}_f = \bm{W}_z [\bm{z}_e, \bm{z}_p]
\end{equation}

The proposed hybrid calibration method effectively improves the quality of LLM knowledge, thereby enhancing the performance of health predictions. By combining insights from both the EHR encoder and the LLM's own generation process, we create a more robust and reliable source of external knowledge for healthcare prediction tasks.

When making predictions, we concatenate the embedding $\bm{h}$ from the existing trained EHR encoder and refined knowledge $\bm{z}_f$ and feed it into a classifier to predict the patient's health status. In IntelliCare, we only update the parameters of the hybrid calibration modules and the predictor during training by the task-specific loss (e.g., cross-entropy for classification) rather than co-training with EHR encoders, which is more flexible and computationally efficient.

\section{Experiments}
\subsection{Experimental Settings}
\subsubsection{Datasets}
We follow previous works \cite{jiang2024graphcare,ma2020adacare,ma2020concare,lu2022context,ma2022patient,xu2023vecocare} to compare models on two large-scale EHR datasets, MIMIC-III \cite{johnson2016mimic} and MIMIC-IV \cite{johnson2023mimic}. We refer to \cite{yang2023pyhealth} for preprocessing. Table \ref{statistic} presents statistics of the processed datasets. These datasets contain many clinical scenarios calling for accurate predictions over diversified clinical signals. We conduct three clinical tasks on these datasets including predicting the \textbf{mortality} at the next visit, \textbf{readmission} prediction, and \textbf{length-of-stay} prediction. Additional details on these datasets can be found in the Appendix.

\begin{table}[h]
    \centering
    \caption{Statistics of the datasets.}
    \small
    \label{statistic}
    \begin{tabular}{lcc}
        \toprule
        \textbf{Dataset} & \textbf{MIMIC-III} & \textbf{MIMIC-IV}\\
        \midrule
        \# Visits & 9,582 & 15,486\\
        \# Conditions per Visit & 13.4 & 14.0\\
        \# Procedures per Visit & 4.7 & 2.5\\
        \# Medications per Visit & 44.8 & 28.8\\
        \# Mortality Rate & 12.1\% & 2.3\%\\
        \# Readmission Rate & 27.9\% & 37.6\%\\
        \# Average Length of Stays & 11.6 & 5.1\\
        \bottomrule
    \end{tabular}
\end{table}

\subsubsection{Evaluation Protocols}
We assess the performance on the binary classification tasks (including mortality and readmission predictions) using the area under the precision-recall curve (AUPRC) and the area under the receiver operating characteristic curve (AUROC). The AUPRC is the most informative and primary evaluation metric when dealing with highly imbalanced and skewed datasets \cite{davis2006relationship} such as healthcare data. The AUROC captures the trade-off between the true positive and false positive rates. For length-of-stay prediction, we follow \cite{jiang2024graphcare} to separate labels into multiple bins and evaluate the AUROC, Kappa, Accuracy, and F1-Score. Kappa \cite{cohen1960coefficient} measures interrater agreement for categorical items to adjust for the level of agreement that would be expected by chance in multiclass classification.
We randomly divide the dataset into a training set containing 80\% of the patients, a validation set of 10\% of patients, and a test set containing the remaining 10\% of instances. The model achieving the best AUROC on the validation set is evaluated on the test set. To mitigate the effects of randomness, we conduct each experiment with three random seeds and report both the mean and standard deviation of the results.

We use LLaMA-3 8B Instruct \cite{meta2024introducing} as the LLM to provide external knowledge, generating eight analyses for each patient. The threshold $\theta$ that selects cohorts is set to 5\%. On a server with four 3090 GPUs, the reasoning of the LLM for all the patients in the MIMIC-III dataset can be completed within 5 hours via vLLM \cite{kwon2023efficient}. We apply gte-large-en-v1.5 \cite{li2023towards} to extract the embeddings of the analyses, which is a 400M BERT-like model supporting context length larger than BERT since BERT may truncate analyses.
Hyperparameters and other implementation details are introduced in the Appendix.

\renewcommand{\dblfloatpagefraction}{.9}
\begin{table*}
    \caption{Performance comparison with standard deviation on the MIMIC-III dataset. The best results are marked in bold.}
    \label{result}
    \centering
    \small
    \setlength{\tabcolsep}{4pt}
    \begin{tabular}{lcccccccc}
        \toprule
        \multirow{3}{*}{\textbf{Model}} & \multicolumn{2}{c}{\textbf{Mortality Prediction}} & \multicolumn{2}{c}{\textbf{Readmission Prediction}} & \multicolumn{4}{c}{\textbf{Length-of-Stay Prediction}}\\
        \cmidrule(lr){2-3} \cmidrule(lr){4-5} \cmidrule(lr){6-9}
        & AUROC(\%) & AUPRC(\%) & AUROC(\%) & AUPRC(\%) & AUROC(\%) & Kappa(\%) & Accuracy(\%) & F1 Score(\%)\\
        \midrule
        Cohort & 59.19 & 14.13 & 58.28 & 34.32 & - & 26.76 & 32.78 & 13.06\\
        \midrule
        StageNet & 58.80{\scriptsize $\pm$2.17} & 15.36{\scriptsize $\pm$0.77} & 58.88{\scriptsize $\pm$1.45} & 36.99{\scriptsize $\pm$0.64} & 77.20{\scriptsize $\pm$0.19} & 31.12{\scriptsize $\pm$0.46} & 36.71{\scriptsize $\pm$0.58} & 22.64{\scriptsize $\pm$1.30} \\
        +IntelliCare & \textbf{67.83{\scriptsize $\pm$0.44}} & \textbf{19.69{\scriptsize $\pm$0.67}} & \textbf{61.54{\scriptsize $\pm$1.58}} & \textbf{39.45{\scriptsize $\pm$1.80}} & \textbf{78.53{\scriptsize $\pm$0.31}} & \textbf{33.60{\scriptsize $\pm$0.39}} & \textbf{39.81{\scriptsize $\pm$0.65}} & \textbf{26.26{\scriptsize $\pm$0.54}} \\
        \midrule
        ConCare & 59.22{\scriptsize $\pm$1.23} & 15.23{\scriptsize $\pm$0.64} & 60.08{\scriptsize $\pm$1.71} & 37.00{\scriptsize $\pm$2.10} & 79.61{\scriptsize $\pm$0.16} & 35.33{\scriptsize $\pm$1.06} & 41.15{\scriptsize $\pm$0.61} & 27.66{\scriptsize $\pm$0.93} \\
        +IntelliCare & \textbf{67.66{\scriptsize $\pm$0.79}} & \textbf{18.85{\scriptsize $\pm$0.47}} & \textbf{63.40{\scriptsize $\pm$0.21}} & \textbf{41.34{\scriptsize $\pm$0.35}} & \textbf{80.19{\scriptsize $\pm$0.09}} & \textbf{36.68{\scriptsize $\pm$0.17}} & \textbf{41.92{\scriptsize $\pm$0.25}} & \textbf{29.23{\scriptsize $\pm$0.31}} \\
        \midrule
        GRASP & 57.29{\scriptsize $\pm$2.42} & 13.99{\scriptsize $\pm$0.68} & 58.76{\scriptsize $\pm$1.48} & 34.92{\scriptsize $\pm$0.89} & 78.95{\scriptsize $\pm$0.70} & 34.81{\scriptsize $\pm$1.58} & 40.36{\scriptsize $\pm$1.03} & 27.15{\scriptsize $\pm$1.76} \\
        +IntelliCare & \textbf{66.88{\scriptsize $\pm$1.55}} & \textbf{18.16{\scriptsize $\pm$0.61}} & \textbf{62.86{\scriptsize $\pm$1.15}} & \textbf{39.28{\scriptsize $\pm$1.58}} & \textbf{79.59{\scriptsize $\pm$0.54}} & \textbf{35.54{\scriptsize $\pm$1.69}} & \textbf{40.89{\scriptsize $\pm$0.92}} & \textbf{28.18{\scriptsize $\pm$1.34}} \\
        \midrule
        SAFARI & 59.47{\scriptsize $\pm$2.26} & 14.74{\scriptsize $\pm$1.26} & 58.36{\scriptsize $\pm$1.12} & 35.36{\scriptsize $\pm$1.06} & 78.15{\scriptsize $\pm$0.71} & 31.02{\scriptsize $\pm$1.59} & 38.48{\scriptsize $\pm$1.06} & 23.52{\scriptsize $\pm$1.91} \\
        +IntelliCare & \textbf{69.94{\scriptsize $\pm$0.36}} & \textbf{19.77{\scriptsize $\pm$0.54}} & \textbf{63.20{\scriptsize $\pm$0.67}} & \textbf{40.89{\scriptsize $\pm$0.34}} & \textbf{78.63{\scriptsize $\pm$0.36}} & \textbf{34.17{\scriptsize $\pm$0.48}} & \textbf{40.12{\scriptsize $\pm$0.32}} & \textbf{26.59{\scriptsize $\pm$0.42}} \\
        \midrule
        GraphCare & 63.65{\scriptsize $\pm$0.25} & 16.02{\scriptsize $\pm$0.30} & 64.02{\scriptsize $\pm$0.22} & 41.19{\scriptsize $\pm$0.13} & 78.42{\scriptsize $\pm$0.26} & \textbf{33.56{\scriptsize $\pm$0.48}} & \textbf{40.78{\scriptsize $\pm$0.36}} & 26.27{\scriptsize $\pm$0.49} \\
        +IntelliCare & \textbf{68.94{\scriptsize $\pm$0.58}} & \textbf{18.78{\scriptsize $\pm$0.27}} & \textbf{66.01{\scriptsize $\pm$0.16}} & \textbf{42.62{\scriptsize $\pm$0.47}} & \textbf{78.58{\scriptsize $\pm$0.10}} & 32.90{\scriptsize $\pm$0.43} & 40.53{\scriptsize $\pm$0.22} & \textbf{26.46{\scriptsize $\pm$0.37}} \\
        \bottomrule
    \end{tabular}
\end{table*}

\begin{table*}
    \caption{Performance comparison with standard deviation on the MIMIC-IV dataset. The best results are marked in bold.}
    \label{result1}
    \centering
    \small
    \setlength{\tabcolsep}{4pt}
    \begin{tabular}{lcccccccc}
        \toprule
        \multirow{3}{*}{\textbf{Model}} & \multicolumn{2}{c}{\textbf{Mortality Prediction}} & \multicolumn{2}{c}{\textbf{Readmission Prediction}} & \multicolumn{4}{c}{\textbf{Length-of-Stay Prediction}}\\
        \cmidrule(lr){2-3} \cmidrule(lr){4-5} \cmidrule(lr){6-9}
        & AUROC(\%) & AUPRC(\%) & AUROC(\%) & AUPRC(\%) & AUROC(\%) & Kappa(\%) & Accuracy(\%) & F1 Score(\%)\\
        \midrule
        Cohort & 65.54 & 4.56 & 64.08 & 53.24 & - & 12.44 & 15.37 & 3.71 \\
        \midrule
        StageNet & 66.44{\scriptsize $\pm$2.06} & 5.74{\scriptsize $\pm$2.16} & 66.05{\scriptsize $\pm$0.28} & 56.44{\scriptsize $\pm$0.33} & 72.49{\scriptsize $\pm$0.21} & 24.08{\scriptsize $\pm$0.99} & 29.26{\scriptsize $\pm$0.82} & \textbf{18.72{\scriptsize $\pm$0.96}} \\
        +IntelliCare & \textbf{74.52{\scriptsize $\pm$1.28}} & \textbf{7.70{\scriptsize $\pm$1.37}} & \textbf{66.89{\scriptsize $\pm$0.09}} & \textbf{57.62{\scriptsize $\pm$0.52}} & \textbf{74.16{\scriptsize $\pm$0.12}} & \textbf{25.07{\scriptsize $\pm$0.49}} & \textbf{29.37{\scriptsize $\pm$0.19}} & 18.68{\scriptsize $\pm$0.36} \\
        \midrule
        ConCare & 69.66{\scriptsize $\pm$1.32} & 6.63{\scriptsize $\pm$0.83} & 67.28{\scriptsize $\pm$0.32} & 57.38{\scriptsize $\pm$0.63} & 73.61{\scriptsize $\pm$0.37} & 24.23{\scriptsize $\pm$0.86} & 29.09{\scriptsize $\pm$0.27} & 18.21{\scriptsize $\pm$0.13} \\
        +IntelliCare & \textbf{77.01{\scriptsize $\pm$0.13}} & \textbf{8.60{\scriptsize $\pm$0.55}} & \textbf{67.80{\scriptsize $\pm$0.34}} & \textbf{58.41{\scriptsize $\pm$0.86}} & \textbf{74.58{\scriptsize $\pm$0.05}} & \textbf{26.07{\scriptsize $\pm$0.52}} & \textbf{30.77{\scriptsize $\pm$0.30}} & \textbf{20.33{\scriptsize $\pm$0.36}} \\
        \midrule
        GRASP & 67.67{\scriptsize $\pm$2.41} & 6.56{\scriptsize $\pm$1.71} & 65.27{\scriptsize $\pm$0.84} & 55.24{\scriptsize $\pm$1.06} & 72.58{\scriptsize $\pm$0.30} & 24.26{\scriptsize $\pm$0.75} & 28.80{\scriptsize $\pm$0.41} & 18.04{\scriptsize $\pm$0.68} \\
        +IntelliCare & \textbf{ 72.80{\scriptsize $\pm$2.55}} & \textbf{7.56{\scriptsize $\pm$1.15}} & \textbf{66.93{\scriptsize $\pm$0.31}} & \textbf{57.57{\scriptsize $\pm$0.38}} & \textbf{74.09{\scriptsize $\pm$0.15}} & \textbf{25.59{\scriptsize $\pm$0.44}} & \textbf{30.03{\scriptsize $\pm$0.58}} & \textbf{19.45{\scriptsize $\pm$0.66}} \\
        \midrule
        SAFARI & 66.88{\scriptsize $\pm$2.01} & 5.55{\scriptsize $\pm$1.03} & 65.44{\scriptsize $\pm$1.11} & 54.73{\scriptsize $\pm$1.21} & 70.50{\scriptsize $\pm$0.48} & 21.75{\scriptsize $\pm$1.75} & 26.35{\scriptsize $\pm$0.43} & 15.11{\scriptsize $\pm$0.59} \\
        +IntelliCare & \textbf{ 77.28{\scriptsize $\pm$1.11}} & \textbf{8.95{\scriptsize $\pm$0.88}} & \textbf{66.70{\scriptsize $\pm$0.17}} & \textbf{56.99{\scriptsize $\pm$0.61}} & \textbf{72.07{\scriptsize $\pm$0.59}} & \textbf{22.41{\scriptsize $\pm$0.26}} & \textbf{27.23{\scriptsize $\pm$0.71}} & \textbf{16.34{\scriptsize $\pm$0.83}} \\
        \midrule
        GraphCare & 72.63{\scriptsize $\pm$0.27} & 7.09{\scriptsize $\pm$0.31} & 67.60{\scriptsize $\pm$0.51} & 58.33{\scriptsize $\pm$0.66} & 72.02{\scriptsize $\pm$0.33} & 22.37{\scriptsize $\pm$0.22} & 28.32{\scriptsize $\pm$0.30} & 17.10{\scriptsize $\pm$0.48} \\
        +IntelliCare & \textbf{76.06{\scriptsize $\pm$0.23}} & \textbf{8.06{\scriptsize $\pm$0.10}} & \textbf{68.12{\scriptsize $\pm$0.28}} & \textbf{58.97{\scriptsize $\pm$0.36}} & \textbf{73.59{\scriptsize $\pm$0.18}} & \textbf{23.50{\scriptsize $\pm$0.33}} & \textbf{29.24{\scriptsize $\pm$0.27}} & \textbf{18.32{\scriptsize $\pm$0.36}} \\
        \bottomrule
    \end{tabular}
\end{table*}

\subsubsection{Baselines} We employ several state-of-the-art models as our baseline models as well as the EHR encoder of IntelliCare, such as \textbf{StageNet} \cite{gao2020stagenet}, which refines the design of long-short-term memory (LSTM) by incorporating personalized disease stage development, and \textbf{ConCare} \cite{ma2020concare}, which consists of a self-attention mechanism to learn feature correlations. We also include works that employ information from other patients, including \textbf{GRASP} \cite{zhang2021grasp}, which exploits similar patients to improve learned representation, and \textbf{SAFARI} \cite{ma2022patient}, which learns feature correlations from a group-wise perspective by a GNN. Besides, we use a model with external knowledge, \textbf{GraphCare} \cite{jiang2024graphcare}, which utilizes knowledge at medical-code level from LLMs and biomedical knowledge graphs to improve encoding. 
Although we have made great efforts in reproducing other models that also leverage external knowledge, such as \cite{ye2021medretriever,lu2022context,xu2023vecocare,yang2023kerprint,xu2024ram}, we are unable to implement them due to the lack of critical details in our settings.

In particular, we report the results of the statistical information for cohorts in section \textit{Patient Cohort Identification}, namely \textbf{Cohort}.

\subsection{Experimental Results}
\subsubsection{Main Results}
We present the overall comparison results on the MIMIC-III and MIMIC-IV datasets in Table \ref{result} and Table \ref{result1}, respectively. We conduct statistical tests and the results show that the improvements are significant. IntelliCare brings significant improvements to existing baselines across most situations. In particular, on the mortality prediction task, IntelliCare achieves results with an average absolute improvement of 7.81\% and 2.92\% on AUROC and AUPRC compared to baseline models, respectively. On the readmission and length-of-stay predictions, our proposed model also brings remarkable improvements. These results demonstrate the effectiveness of IntelliCare in assisting existing EHR models by incorporating knowledge from LLMs.

By comparing the results among tasks, we observe that IntelliCare performs better on the mortality prediction task than on the other two tasks. This is because existing models may not adequately capture the complex dependencies and intricate relationships associated with the mortality prediction task, which can be effectively addressed by introducing external knowledge. Although GraphCare leverages knowledge from LLMs and knowledge graphs, IntelliCare can still boost its performance, demonstrating the generality of our proposed method. We note that the gains brought by IntelliCare may differ across baseline models, which can be attributed to the baseline model structure and information fusion strategy. Applying better fusion approaches in the predictor has the potential to further improve performance.

In particular, using only our unsupervised clustering method cannot achieve performance that exceeds that of existing supervised methods, indicating that the main improvement of IntelliCare comes from external knowledge injection rather than cluster statistics. These results also show that the performance of IntelliCare is more stable than the baselines, as indicated by the smaller standard deviation.

\subsubsection{Ablation Study}
To evaluate the effectiveness of each component in IntelliCare, we use two outstanding methods, ConCare and SAFARI, as EHR encoders and perform ablation studies on the MIMIC-III dataset. We introduce four variants to investigate the benefits of every design on model performance as follows:
\begin{itemize}[leftmargin=*]
    \item Cohort Only combines the embedding of textual cohort information with the EHR encoder to predict the target.
    \item IntelliCare$_{t-}$ removes the cohort prompt and task-relevant instructions in trigger prompt when generating knowledge, allowing the LLMs to only summarize the patient's health status without analyzing with respect to the task.
    \item IntelliCare$_{c-}$ removes the cohort prompt, which merely utilizes the internal knowledge in LLMs when conducting analyses on the patients.
    \item IntelliCare$_{p-}$ removes the perplexity-guided weighting mechanism. It only uses the information from the EHR encoder to adjust the analyses from the LLMs.
    \item IntelliCare$_{r-}$ removes the branch of the encoder-driven rectification in hybrid analysis refinement. It refines the knowledge from LLMs by perplexity exclusively.
\end{itemize}

\begin{table*}
    \caption{Ablation study of IntelliCare on the MIMIC-III dataset.}
    \label{ablation}
    \centering
    \small
    \setlength{\tabcolsep}{4pt}
    \begin{tabular}{lcccccccc}
        \toprule
        \multirow{3}{*}{\textbf{Model}} & \multicolumn{2}{c}{\textbf{Mortality Prediction}} & \multicolumn{2}{c}{\textbf{Readmission Prediction}} & \multicolumn{4}{c}{\textbf{Length-of-Stay Prediction}}\\
        \cmidrule(lr){2-3} \cmidrule(lr){4-5} \cmidrule(lr){6-9}
        & AUROC(\%) & AUPRC(\%) & AUROC(\%) & AUPRC(\%) & AUROC(\%) & Kappa(\%) & Accuracy(\%) & F1 Score(\%)\\
        \midrule
        ConCare & 59.22{\scriptsize $\pm$1.23} & 15.23{\scriptsize $\pm$0.64} & 60.08{\scriptsize $\pm$1.71} & 37.00{\scriptsize $\pm$2.10} & 79.61{\scriptsize $\pm$0.16} & 35.33{\scriptsize $\pm$1.06} & 41.15{\scriptsize $\pm$0.61} & 27.66{\scriptsize $\pm$0.93} \\
        +Cohort Only & 59.66{\scriptsize $\pm$2.41} & 15.24{\scriptsize $\pm$0.74} & 59.52{\scriptsize $\pm$1.40} & 36.91{\scriptsize $\pm$1.78} & 79.76{\scriptsize $\pm$0.06} & 35.38{\scriptsize $\pm$0.79} & 41.31{\scriptsize $\pm$0.51} & 28.32{\scriptsize $\pm$0.95} \\
        +IntelliCare$_{t-}$ & 66.07{\scriptsize $\pm$1.32} & 18.11{\scriptsize $\pm$0.19} & 62.92{\scriptsize $\pm$0.74} & 40.30{\scriptsize $\pm$0.99} & 79.65{\scriptsize $\pm$0.06} & 36.02{\scriptsize $\pm$0.99} & 41.37{\scriptsize $\pm$0.57} & 28.59{\scriptsize $\pm$0.92} \\
        +IntelliCare$_{c-}$ & 66.90{\scriptsize $\pm$0.89} & 18.49{\scriptsize $\pm$0.36} & 63.07{\scriptsize $\pm$0.81} & 40.88{\scriptsize $\pm$1.01} & 79.86{\scriptsize $\pm$0.06} & 35.81{\scriptsize $\pm$0.73} & 41.73{\scriptsize $\pm$0.46} & 28.62{\scriptsize $\pm$0.65} \\
        +IntelliCare$_{p-}$ & 67.08{\scriptsize $\pm$1.21} & 18.66{\scriptsize $\pm$0.63} & 62.54{\scriptsize $\pm$1.16} & 40.35{\scriptsize $\pm$1.39} & 80.06{\scriptsize $\pm$0.14} & 35.17{\scriptsize $\pm$0.28} & 41.18{\scriptsize $\pm$0.38} & 28.23{\scriptsize $\pm$0.37} \\
        +IntelliCare$_{r-}$ & 67.46{\scriptsize $\pm$0.56} & 18.83{\scriptsize $\pm$0.57} & 63.27{\scriptsize $\pm$1.01} & 41.10{\scriptsize $\pm$0.88} & 79.74{\scriptsize $\pm$0.10} & 35.33{\scriptsize $\pm$0.10} & 41.21{\scriptsize $\pm$0.27} & 27.95{\scriptsize $\pm$0.13} \\
        +IntelliCare & \textbf{67.66{\scriptsize $\pm$0.79}} & \textbf{18.85{\scriptsize $\pm$0.47}} & \textbf{63.40{\scriptsize $\pm$0.21}} & \textbf{41.34{\scriptsize $\pm$0.35}} & \textbf{80.19{\scriptsize $\pm$0.09}} & \textbf{36.68{\scriptsize $\pm$0.17}} & \textbf{41.92{\scriptsize $\pm$0.25}} & \textbf{29.23{\scriptsize $\pm$0.31}} \\
        \midrule
        SAFARI & 59.47{\scriptsize $\pm$2.26} & 14.74{\scriptsize $\pm$1.26} & 58.36{\scriptsize $\pm$1.12} & 35.36{\scriptsize $\pm$1.06} & 78.15{\scriptsize $\pm$0.71} & 31.02{\scriptsize $\pm$1.59} & 38.48{\scriptsize $\pm$1.06} & 23.52{\scriptsize $\pm$1.91} \\
        +Cohort & 59.17{\scriptsize $\pm$1.08} & 14.35{\scriptsize $\pm$0.52} & 62.92{\scriptsize $\pm$0.74} & 39.30{\scriptsize $\pm$0.99} & 78.65{\scriptsize $\pm$0.06} & \textbf{35.02{\scriptsize $\pm$0.99}} & \textbf{40.37{\scriptsize $\pm$0.57}} & 25.29{\scriptsize $\pm$0.92} \\
        +IntelliCare$_{c-}$ & 69.08{\scriptsize $\pm$0.28} & 19.55{\scriptsize $\pm$0.34} & 62.17{\scriptsize $\pm$0.58} & 38.95{\scriptsize $\pm$0.67} & \textbf{78.80{\scriptsize $\pm$0.52}} & 33.54{\scriptsize $\pm$0.36} & 39.13{\scriptsize $\pm$0.84} & 25.41{\scriptsize $\pm$0.58} \\
        +IntelliCare$_{c-}$ & 69.08{\scriptsize $\pm$0.28} & 19.55{\scriptsize $\pm$0.34} & 62.17{\scriptsize $\pm$0.58} & 38.95{\scriptsize $\pm$0.67} & 78.80{\scriptsize $\pm$0.52} & 33.54{\scriptsize $\pm$0.36} & 39.13{\scriptsize $\pm$0.84} & 25.41{\scriptsize $\pm$0.58} \\
        +IntelliCare$_{p-}$ & 69.53{\scriptsize $\pm$0.73} & 19.45{\scriptsize $\pm$0.60} & 63.06{\scriptsize $\pm$0.97} & 40.86{\scriptsize $\pm$0.34} & 78.58{\scriptsize $\pm$0.48} & 33.80{\scriptsize $\pm$0.82} & 40.18{\scriptsize $\pm$0.59} & 26.36{\scriptsize $\pm$0.81} \\
        +IntelliCare$_{r-}$ & 69.47{\scriptsize $\pm$0.77} & 19.67{\scriptsize $\pm$0.89} & 62.82{\scriptsize $\pm$0.96} & 39.32{\scriptsize $\pm$1.17} & 78.33{\scriptsize $\pm$0.59} & 32.64{\scriptsize $\pm$0.61} & 39.10{\scriptsize $\pm$0.98} & 25.45{\scriptsize $\pm$1.28} \\
        +IntelliCare & \textbf{69.94{\scriptsize $\pm$0.36}} & \textbf{19.77{\scriptsize $\pm$0.54}} & \textbf{63.20{\scriptsize $\pm$0.67}} & \textbf{40.89{\scriptsize $\pm$0.34}} & 78.63{\scriptsize $\pm$0.36} & 34.17{\scriptsize $\pm$0.48} & 40.12{\scriptsize $\pm$0.32} & \textbf{26.59{\scriptsize $\pm$0.42}} \\
        \bottomrule
    \end{tabular}
\end{table*}

\begin{figure*}[h]
    \centering
    \includegraphics[width=1.9\columnwidth]{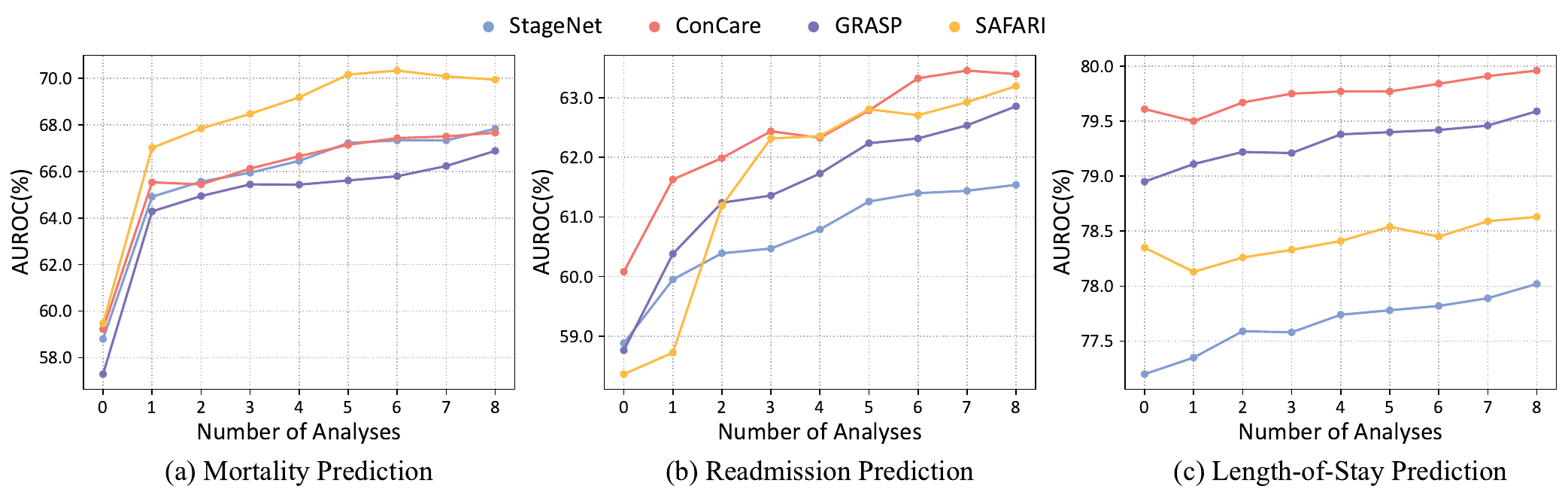}
    \caption{Ablation comparison of IntelliCare for different numbers of analyses on the MIMIC-III dataset.}
    \label{k-mimiciii}
    \vspace{-0.25cm}
\end{figure*}

As shown in Table \ref{ablation}, we observe that all components and designs contribute to the improvement. Merely use cohort information cannot bring improvements, demonstrating that the enhancement comes from the knowledge of the LLM. Combining task analysis yields improvements in most settings, validating the ability of LLMs to provide specific insights into different clinical tasks. Without statistical information from cohorts, the performance of IntelliCare$_{c-}$ decays compared to IntelliCare, verifying that the introduction of cohort information can effectively reduce the variance of LLM outputs and improve the quality of the knowledge it provides. Notably, the encoder-driven rectification and the perplexity-guided weighting both bring benefits, but the extent of the gains may vary across EHR encoders and tasks. Combining them can leverage their respective advantages and achieve better results. These findings highlight the necessity of novel designs to mitigate the variance of LLMs in the inference process and knowledge fusion strategies.

\subsubsection{Analysis on the Number of LLM Analyses}
To further investigate the impact of the number of analyses, we conduct a comprehensive experiment on the MIMIC-III dataset by varying the number of analyses in integrating hybrid analysis refinement. Since the training of GraphCare consumes too much time, we only conduct this experiment on the other EHR encoders. The results of AUROC on the three tasks are shown in Figure \ref{k-mimiciii}. Zero analysis means that we only use the EHR encoder to predict the target, and the performance is the same as the baseline model in Table \ref{result}. As illustrated, the performance of IntelliCare increases with the number of analyses. Though the improvement becomes less significant when the number of analyses exceeds three, more analyses still bring better performance. This indicates that the model can benefit from more analyses to a certain extent, but the boosting on the performance will be limited when the number of analyses is sufficient. We suggest that the number of LLM analyses can be decided based on the computational resources in real-world application scenarios.

\section{Conclusions}
In this paper, we propose IntelliCare, a novel framework that leverages external knowledge from LLMs to enhance existing EHR models. To control the variance of LLMs and effectively integrate their knowledge into EHR models, we introduce patient cohort identification and hybrid analysis refinement. Specifically, we design a method to identify patient cohorts and utilize the statistical information of these cohorts to compensate for the lack of knowledge of LLM about clinical tasks. We then generate multiple analyses and combine them through two perspectives, information from EHR models and the perplexities of the generating process, to reduce the inconsistent knowledge from LLMs.
We conduct extensive experiments on three clinical tasks on two real-world datasets using recently popular models to evaluate the effectiveness of novel designs in IntelliCare. The experimental results demonstrate that IntelliCare consistently brings significant performance and stability improvements over them. 
We hope that our method will contribute to improving the healthcare system and assisting physicians.

\bibliography{reference}

\begin{thebibliography}{58}
\providecommand{\natexlab}[1]{#1}

\bibitem[{Ashfaq et~al.(2019)Ashfaq, Sant’Anna, Lingman, and
  Nowaczyk}]{ashfaq2019readmission}
Ashfaq, A.; Sant’Anna, A.; Lingman, M.; and Nowaczyk, S. 2019.
\newblock Readmission prediction using deep learning on electronic health
  records.
\newblock \emph{Journal of biomedical informatics}, 97: 103256.

\bibitem[{Ba, Kiros, and Hinton(2016)}]{ba2016layer}
Ba, J.~L.; Kiros, J.~R.; and Hinton, G.~E. 2016.
\newblock Layer normalization.
\newblock \emph{arXiv preprint arXiv:1607.06450}.

\bibitem[{Bai et~al.(2018)Bai, Zhang, Egleston, and
  Vucetic}]{bai2018interpretable}
Bai, T.; Zhang, S.; Egleston, B.~L.; and Vucetic, S. 2018.
\newblock Interpretable representation learning for healthcare via capturing
  disease progression through time.
\newblock In \emph{Proceedings of the 24th ACM SIGKDD international conference
  on knowledge discovery \& data mining}, 43--51.

\bibitem[{Baytas et~al.(2017)Baytas, Xiao, Zhang, Wang, Jain, and
  Zhou}]{baytas2017patient}
Baytas, I.~M.; Xiao, C.; Zhang, X.; Wang, F.; Jain, A.~K.; and Zhou, J. 2017.
\newblock Patient subtyping via time-aware LSTM networks.
\newblock In \emph{Proceedings of the 23rd ACM SIGKDD international conference
  on knowledge discovery and data mining}, 65--74.

\bibitem[{Bishop and Nasrabadi(2006)}]{bishop2006pattern}
Bishop, C.~M.; and Nasrabadi, N.~M. 2006.
\newblock \emph{Pattern recognition and machine learning}, volume~4.
\newblock Springer.

\bibitem[{Cai et~al.(2022)Cai, Sun, Song, Zhang, Hong, and
  Li}]{cai2022hypergraph}
Cai, D.; Sun, C.; Song, M.; Zhang, B.; Hong, S.; and Li, H. 2022.
\newblock Hypergraph contrastive learning for electronic health records.
\newblock In \emph{Proceedings of the 2022 SIAM International Conference on
  Data Mining (SDM)}, 127--135. SIAM.

\bibitem[{Choi et~al.(2016{\natexlab{a}})Choi, Bahadori, Searles, Coffey,
  Thompson, Bost, Tejedor-Sojo, and Sun}]{choi2016multi}
Choi, E.; Bahadori, M.~T.; Searles, E.; Coffey, C.; Thompson, M.; Bost, J.;
  Tejedor-Sojo, J.; and Sun, J. 2016{\natexlab{a}}.
\newblock Multi-layer representation learning for medical concepts.
\newblock In \emph{proceedings of the 22nd ACM SIGKDD international conference
  on knowledge discovery and data mining}, 1495--1504.

\bibitem[{Choi et~al.(2017)Choi, Bahadori, Song, Stewart, and
  Sun}]{choi2017gram}
Choi, E.; Bahadori, M.~T.; Song, L.; Stewart, W.~F.; and Sun, J. 2017.
\newblock GRAM: graph-based attention model for healthcare representation
  learning.
\newblock In \emph{Proceedings of the 23rd ACM SIGKDD international conference
  on knowledge discovery and data mining}, 787--795.

\bibitem[{Choi et~al.(2016{\natexlab{b}})Choi, Bahadori, Sun, Kulas, Schuetz,
  and Stewart}]{choi2016retain}
Choi, E.; Bahadori, M.~T.; Sun, J.; Kulas, J.; Schuetz, A.; and Stewart, W.
  2016{\natexlab{b}}.
\newblock Retain: An interpretable predictive model for healthcare using
  reverse time attention mechanism.
\newblock \emph{Advances in neural information processing systems}, 29.

\bibitem[{Choi et~al.(2018)Choi, Xiao, Stewart, and Sun}]{choi2018mime}
Choi, E.; Xiao, C.; Stewart, W.; and Sun, J. 2018.
\newblock Mime: Multilevel medical embedding of electronic health records for
  predictive healthcare.
\newblock \emph{Advances in neural information processing systems}, 31.

\bibitem[{Choi et~al.(2020)Choi, Xu, Li, Dusenberry, Flores, Xue, and
  Dai}]{choi2020learning}
Choi, E.; Xu, Z.; Li, Y.; Dusenberry, M.; Flores, G.; Xue, E.; and Dai, A.
  2020.
\newblock Learning the graphical structure of electronic health records with
  graph convolutional transformer.
\newblock In \emph{Proceedings of the AAAI conference on artificial
  intelligence}, volume~34, 606--613.

\bibitem[{Cohen(1960)}]{cohen1960coefficient}
Cohen, J. 1960.
\newblock A coefficient of agreement for nominal scales.
\newblock \emph{Educational and psychological measurement}, 20(1): 37--46.

\bibitem[{Corduneanu and Bishop(2001)}]{corduneanu2001variational}
Corduneanu, A.; and Bishop, C.~M. 2001.
\newblock Variational Bayesian model selection for mixture distributions.
\newblock In \emph{Artificial intelligence and Statistics}, volume 2001,
  27--34. Morgan Kaufmann Waltham, MA.

\bibitem[{Cui et~al.(2024)Cui, Shen, Zhang, Shao, Qin, Ho, and
  Yang}]{cui2024llms}
Cui, H.; Shen, Z.; Zhang, J.; Shao, H.; Qin, L.; Ho, J.~C.; and Yang, C. 2024.
\newblock LLMs-based Few-Shot Disease Predictions using EHR: A Novel Approach
  Combining Predictive Agent Reasoning and Critical Agent Instruction.
\newblock \emph{arXiv preprint arXiv:2403.15464}.

\bibitem[{Davis and Goadrich(2006)}]{davis2006relationship}
Davis, J.; and Goadrich, M. 2006.
\newblock The relationship between Precision-Recall and ROC curves.
\newblock In \emph{Proceedings of the 23rd international conference on Machine
  learning}, 233--240.

\bibitem[{Dice(1945)}]{dice1945measures}
Dice, L.~R. 1945.
\newblock Measures of the amount of ecologic association between species.
\newblock \emph{Ecology}, 26(3): 297--302.

\bibitem[{Dong et~al.(2024)Dong, Xu, Kong, Sui, and Li}]{dong2024statistical}
Dong, Q.; Xu, J.; Kong, L.; Sui, Z.; and Li, L. 2024.
\newblock Statistical Knowledge Assessment for Large Language Models.
\newblock \emph{Advances in Neural Information Processing Systems}, 36.

\bibitem[{Gao et~al.(2020)Gao, Xiao, Wang, Tang, Glass, and
  Sun}]{gao2020stagenet}
Gao, J.; Xiao, C.; Wang, Y.; Tang, W.; Glass, L.~M.; and Sun, J. 2020.
\newblock Stagenet: Stage-aware neural networks for health risk prediction.
\newblock In \emph{Proceedings of The Web Conference 2020}, 530--540.

\bibitem[{Gao et~al.(2022)Gao, Yang, Heintz, Barrows, Albers, Stapel, Warfield,
  Cross, and Sun}]{gao2022medml}
Gao, J.; Yang, C.; Heintz, J.; Barrows, S.; Albers, E.; Stapel, M.; Warfield,
  S.; Cross, A.; and Sun, J. 2022.
\newblock MedML: fusing medical knowledge and machine learning models for early
  pediatric COVID-19 hospitalization and severity prediction.
\newblock \emph{Iscience}, 25(9).

\bibitem[{Jelinek et~al.(1977)Jelinek, Mercer, Bahl, and
  Baker}]{jelinek1977perplexity}
Jelinek, F.; Mercer, R.~L.; Bahl, L.~R.; and Baker, J.~K. 1977.
\newblock Perplexity—a measure of the difficulty of speech recognition tasks.
\newblock \emph{The Journal of the Acoustical Society of America}, 62(S1):
  S63--S63.

\bibitem[{Jiang et~al.(2024)Jiang, Xiao, Cross, and Sun}]{jiang2024graphcare}
Jiang, P.; Xiao, C.; Cross, A.~R.; and Sun, J. 2024.
\newblock GraphCare: Enhancing Healthcare Predictions with Personalized
  Knowledge Graphs.
\newblock In \emph{The Twelfth International Conference on Learning
  Representations}.

\bibitem[{Johnson et~al.(2023)Johnson, Bulgarelli, Shen, Gayles, Shammout,
  Horng, Pollard, Hao, Moody, Gow et~al.}]{johnson2023mimic}
Johnson, A.~E.; Bulgarelli, L.; Shen, L.; Gayles, A.; Shammout, A.; Horng, S.;
  Pollard, T.~J.; Hao, S.; Moody, B.; Gow, B.; et~al. 2023.
\newblock MIMIC-IV, a freely accessible electronic health record dataset.
\newblock \emph{Scientific data}, 10(1): 1.

\bibitem[{Johnson et~al.(2016)Johnson, Pollard, Shen, Lehman, Feng, Ghassemi,
  Moody, Szolovits, Anthony~Celi, and Mark}]{johnson2016mimic}
Johnson, A.~E.; Pollard, T.~J.; Shen, L.; Lehman, L.-w.~H.; Feng, M.; Ghassemi,
  M.; Moody, B.; Szolovits, P.; Anthony~Celi, L.; and Mark, R.~G. 2016.
\newblock MIMIC-III, a freely accessible critical care database.
\newblock \emph{Scientific data}, 3(1): 1--9.

\bibitem[{Kingma and Ba(2014)}]{kingma2014adam}
Kingma, D.~P.; and Ba, J. 2014.
\newblock Adam: A method for stochastic optimization.
\newblock \emph{arXiv preprint arXiv:1412.6980}.

\bibitem[{Kwon et~al.(2023)Kwon, Li, Zhuang, Sheng, Zheng, Yu, Gonzalez, Zhang,
  and Stoica}]{kwon2023efficient}
Kwon, W.; Li, Z.; Zhuang, S.; Sheng, Y.; Zheng, L.; Yu, C.~H.; Gonzalez, J.;
  Zhang, H.; and Stoica, I. 2023.
\newblock Efficient memory management for large language model serving with
  pagedattention.
\newblock In \emph{Proceedings of the 29th Symposium on Operating Systems
  Principles}, 611--626.

\bibitem[{Li et~al.(2023)Li, Zhang, Zhang, Long, Xie, and
  Zhang}]{li2023towards}
Li, Z.; Zhang, X.; Zhang, Y.; Long, D.; Xie, P.; and Zhang, M. 2023.
\newblock Towards general text embeddings with multi-stage contrastive
  learning.
\newblock \emph{arXiv preprint arXiv:2308.03281}.

\bibitem[{Lu, Han, and Ning(2022)}]{lu2022context}
Lu, C.; Han, T.; and Ning, Y. 2022.
\newblock Context-aware health event prediction via transition functions on
  dynamic disease graphs.
\newblock In \emph{Proceedings of the AAAI Conference on Artificial
  Intelligence}, volume~36, 4567--4574.

\bibitem[{Lu et~al.(2021)Lu, Reddy, Chakraborty, Kleinberg, and
  Ning}]{lu2021collaborative}
Lu, C.; Reddy, C.~K.; Chakraborty, P.; Kleinberg, S.; and Ning, Y. 2021.
\newblock Collaborative Graph Learning with Auxiliary Text for Temporal Event
  Prediction in Healthcare.
\newblock \emph{International Joint Conference on Artificial Intelligence}.

\bibitem[{Luo et~al.(2020)Luo, Ye, Xiao, and Ma}]{luo2020hitanet}
Luo, J.; Ye, M.; Xiao, C.; and Ma, F. 2020.
\newblock Hitanet: Hierarchical time-aware attention networks for risk
  prediction on electronic health records.
\newblock In \emph{Proceedings of the 26th ACM SIGKDD International Conference
  on Knowledge Discovery \& Data Mining}, 647--656.

\bibitem[{Ma et~al.(2018)Ma, You, Xiao, Chitta, Zhou, and Gao}]{ma2018kame}
Ma, F.; You, Q.; Xiao, H.; Chitta, R.; Zhou, J.; and Gao, J. 2018.
\newblock Kame: Knowledge-based attention model for diagnosis prediction in
  healthcare.
\newblock In \emph{Proceedings of the 27th ACM international conference on
  information and knowledge management}, 743--752.

\bibitem[{Ma et~al.(2020{\natexlab{a}})Ma, Gao, Wang, Zhang, Wang, Ruan, Tang,
  Gao, and Ma}]{ma2020adacare}
Ma, L.; Gao, J.; Wang, Y.; Zhang, C.; Wang, J.; Ruan, W.; Tang, W.; Gao, X.;
  and Ma, X. 2020{\natexlab{a}}.
\newblock Adacare: Explainable clinical health status representation learning
  via scale-adaptive feature extraction and recalibration.
\newblock In \emph{Proceedings of the AAAI Conference on Artificial
  Intelligence}, volume~34, 825--832.

\bibitem[{Ma et~al.(2021)Ma, Ma, Gao, Jiao, Yu, Zhang, Ruan, Wang, Tang, and
  Wang}]{ma2021distilling}
Ma, L.; Ma, X.; Gao, J.; Jiao, X.; Yu, Z.; Zhang, C.; Ruan, W.; Wang, Y.; Tang,
  W.; and Wang, J. 2021.
\newblock Distilling knowledge from publicly available online EMR data to
  emerging epidemic for prognosis.
\newblock In \emph{Proceedings of the Web Conference 2021}, 3558--3568.

\bibitem[{Ma et~al.(2023)Ma, Zhang, Gao, Jiao, Yu, Zhu, Wang, Ma, Wang, Tang
  et~al.}]{ma2023mortality}
Ma, L.; Zhang, C.; Gao, J.; Jiao, X.; Yu, Z.; Zhu, Y.; Wang, T.; Ma, X.; Wang,
  Y.; Tang, W.; et~al. 2023.
\newblock Mortality prediction with adaptive feature importance recalibration
  for peritoneal dialysis patients.
\newblock \emph{Patterns}, 4(12).

\bibitem[{Ma et~al.(2020{\natexlab{b}})Ma, Zhang, Wang, Ruan, Wang, Tang, Ma,
  Gao, and Gao}]{ma2020concare}
Ma, L.; Zhang, C.; Wang, Y.; Ruan, W.; Wang, J.; Tang, W.; Ma, X.; Gao, X.; and
  Gao, J. 2020{\natexlab{b}}.
\newblock Concare: Personalized clinical feature embedding via capturing the
  healthcare context.
\newblock In \emph{Proceedings of the AAAI Conference on Artificial
  Intelligence}, volume~34, 833--840.

\bibitem[{Ma et~al.(2022)Ma, Wang, Chu, Ma, Tang, Zhao, Yuan, and
  Wang}]{ma2022patient}
Ma, X.; Wang, Y.; Chu, X.; Ma, L.; Tang, W.; Zhao, J.; Yuan, Y.; and Wang, G.
  2022.
\newblock Patient health representation learning via correlational sparse prior
  of medical features.
\newblock \emph{IEEE Transactions on Knowledge and Data Engineering}.

\bibitem[{McInnes et~al.(2018)McInnes, Healy, Saul, and
  Gro{\ss}berger}]{mcinnes2018umap}
McInnes, L.; Healy, J.; Saul, N.; and Gro{\ss}berger, L. 2018.
\newblock UMAP: Uniform Manifold Approximation and Projection.
\newblock \emph{Journal of Open Source Software}, 3(29): 861.

\bibitem[{Meta(2024)}]{meta2024introducing}
Meta, A. 2024.
\newblock Introducing meta llama 3: The most capable openly available llm to
  date.
\newblock \emph{Meta AI}.

\bibitem[{Shoham and Rappoport(2023)}]{shoham2023cpllm}
Shoham, O.~B.; and Rappoport, N. 2023.
\newblock Cpllm: Clinical prediction with large language models.
\newblock \emph{arXiv preprint arXiv:2309.11295}.

\bibitem[{Song et~al.(2018)Song, Rajan, Thiagarajan, and
  Spanias}]{song2018attend}
Song, H.; Rajan, D.; Thiagarajan, J.; and Spanias, A. 2018.
\newblock Attend and diagnose: Clinical time series analysis using attention
  models.
\newblock In \emph{Proceedings of the AAAI conference on artificial
  intelligence}, volume~32.

\bibitem[{Sorensen(1948)}]{sorensen1948method}
Sorensen, T. 1948.
\newblock A method of establishing groups of equal amplitude in plant sociology
  based on similarity of species content and its application to analyses of the
  vegetation on Danish commons.
\newblock \emph{Biologiske skrifter}, 5: 1--34.

\bibitem[{Tan et~al.(2022)Tan, Yang, Wei, Chen, Liu, Li, Zhou, and
  Zheng}]{tan2022metacare}
Tan, Y.; Yang, C.; Wei, X.; Chen, C.; Liu, W.; Li, L.; Zhou, J.; and Zheng, X.
  2022.
\newblock Metacare++: Meta-learning with hierarchical subtyping for cold-start
  diagnosis prediction in healthcare data.
\newblock In \emph{Proceedings of the 45th International ACM SIGIR Conference
  on Research and Development in Information Retrieval}, 449--459.

\bibitem[{Vaswani et~al.(2017)Vaswani, Shazeer, Parmar, Uszkoreit, Jones,
  Gomez, Kaiser, and Polosukhin}]{vaswani2017attention}
Vaswani, A.; Shazeer, N.; Parmar, N.; Uszkoreit, J.; Jones, L.; Gomez, A.~N.;
  Kaiser, {\L}.; and Polosukhin, I. 2017.
\newblock Attention is all you need.
\newblock \emph{Advances in neural information processing systems}, 30.

\bibitem[{Wang et~al.(2024)Wang, Ahn, Dalal, Zhang, Pan, Zhang, Chen, Dodge,
  Wang, and Zhou}]{wang2024augmented}
Wang, J.; Ahn, S.; Dalal, T.; Zhang, X.; Pan, W.; Zhang, Q.; Chen, B.; Dodge,
  H.~H.; Wang, F.; and Zhou, J. 2024.
\newblock Augmented Risk Prediction for the Onset of Alzheimer's Disease from
  Electronic Health Records with Large Language Models.
\newblock \emph{arXiv preprint arXiv:2405.16413}.

\bibitem[{Xu et~al.(2024{\natexlab{a}})Xu, Qi, Wang, Wang, Zhang, and
  Xu}]{xu2024knowledge}
Xu, R.; Qi, Z.; Wang, C.; Wang, H.; Zhang, Y.; and Xu, W. 2024{\natexlab{a}}.
\newblock Knowledge Conflicts for LLMs: A Survey.
\newblock \emph{arXiv preprint arXiv:2403.08319}.

\bibitem[{Xu et~al.(2024{\natexlab{b}})Xu, Shi, Yu, Zhuang, Jin, Wang, Ho, and
  Yang}]{xu2024ram}
Xu, R.; Shi, W.; Yu, Y.; Zhuang, Y.; Jin, B.; Wang, M.~D.; Ho, J.~C.; and Yang,
  C. 2024{\natexlab{b}}.
\newblock RAM-EHR: Retrieval Augmentation Meets Clinical Predictions on
  Electronic Health Records.
\newblock In \emph{Proceedings of the 62nd Annual Meeting of the Association
  for Computational Linguistics}.

\bibitem[{Xu et~al.(2022)Xu, Yu, Zhang, Ali, Ho, and
  Yang}]{xu2022counterfactual}
Xu, R.; Yu, Y.; Zhang, C.; Ali, M.~K.; Ho, J.~C.; and Yang, C. 2022.
\newblock Counterfactual and Factual Reasoning over Hypergraphs for
  Interpretable Clinical Predictions on EHR.
\newblock In \emph{Machine Learning for Health}, 259--278. PMLR.

\bibitem[{Xu et~al.(2023{\natexlab{a}})Xu, Chu, Yang, Wang, Zou, Ding, Zhao,
  Wang, and Xie}]{xu2023seqcare}
Xu, Y.; Chu, X.; Yang, K.; Wang, Z.; Zou, P.; Ding, H.; Zhao, J.; Wang, Y.; and
  Xie, B. 2023{\natexlab{a}}.
\newblock Seqcare: Sequential training with external medical knowledge graph
  for diagnosis prediction in healthcare data.
\newblock In \emph{Proceedings of the ACM Web Conference 2023}, 2819--2830.

\bibitem[{Xu et~al.(2023{\natexlab{b}})Xu, Yang, Zhang, Zou, Wang, Ding, Zhao,
  Wang, and Xie}]{xu2023vecocare}
Xu, Y.; Yang, K.; Zhang, C.; Zou, P.; Wang, Z.; Ding, H.; Zhao, J.; Wang, Y.;
  and Xie, B. 2023{\natexlab{b}}.
\newblock VecoCare: visit sequences-clinical notes joint learning for diagnosis
  prediction in healthcare data.
\newblock In \emph{Proceedings of the Thirty-Second International Joint
  Conference on Artificial Intelligence, IJCAI-23}, 4921--4929.

\bibitem[{Yang et~al.(2023{\natexlab{a}})Yang, Wu, Jiang, Lin, Gao, Danek, and
  Sun}]{yang2023pyhealth}
Yang, C.; Wu, Z.; Jiang, P.; Lin, Z.; Gao, J.; Danek, B.~P.; and Sun, J.
  2023{\natexlab{a}}.
\newblock Pyhealth: A deep learning toolkit for healthcare applications.
\newblock In \emph{Proceedings of the 29th ACM SIGKDD Conference on Knowledge
  Discovery and Data Mining}, 5788--5789.

\bibitem[{Yang et~al.(2023{\natexlab{b}})Yang, Xu, Zou, Ding, Zhao, Wang, and
  Xie}]{yang2023kerprint}
Yang, K.; Xu, Y.; Zou, P.; Ding, H.; Zhao, J.; Wang, Y.; and Xie, B.
  2023{\natexlab{b}}.
\newblock KerPrint: local-global knowledge graph enhanced diagnosis prediction
  for retrospective and prospective interpretations.
\newblock In \emph{Proceedings of the AAAI Conference on Artificial
  Intelligence}, volume~37, 5357--5365.

\bibitem[{Ye et~al.(2021{\natexlab{a}})Ye, Cui, Wang, Luo, Xiao, and
  Ma}]{ye2021medpath}
Ye, M.; Cui, S.; Wang, Y.; Luo, J.; Xiao, C.; and Ma, F. 2021{\natexlab{a}}.
\newblock Medpath: Augmenting health risk prediction via medical knowledge
  paths.
\newblock In \emph{Proceedings of the Web Conference 2021}, 1397--1409.

\bibitem[{Ye et~al.(2021{\natexlab{b}})Ye, Cui, Wang, Luo, Xiao, and
  Ma}]{ye2021medretriever}
Ye, M.; Cui, S.; Wang, Y.; Luo, J.; Xiao, C.; and Ma, F. 2021{\natexlab{b}}.
\newblock Medretriever: Target-driven interpretable health risk prediction via
  retrieving unstructured medical text.
\newblock In \emph{Proceedings of the 30th ACM International Conference on
  Information \& Knowledge Management}, 2414--2423.

\bibitem[{Ye et~al.(2020)Ye, Luo, Xiao, and Ma}]{ye2020lsan}
Ye, M.; Luo, J.; Xiao, C.; and Ma, F. 2020.
\newblock Lsan: Modeling long-term dependencies and short-term correlations
  with hierarchical attention for risk prediction.
\newblock In \emph{Proceedings of the 29th ACM International Conference on
  Information \& Knowledge Management}, 1753--1762.

\bibitem[{Yin et~al.(2019)Yin, Zhao, Qian, Lv, and Zhang}]{yin2019domain}
Yin, C.; Zhao, R.; Qian, B.; Lv, X.; and Zhang, P. 2019.
\newblock Domain knowledge guided deep learning with electronic health records.
\newblock In \emph{2019 IEEE International Conference on Data Mining (ICDM)},
  738--747. IEEE.

\bibitem[{Zhang et~al.(2021)Zhang, Gao, Ma, Wang, Wang, and
  Tang}]{zhang2021grasp}
Zhang, C.; Gao, X.; Ma, L.; Wang, Y.; Wang, J.; and Tang, W. 2021.
\newblock GRASP: generic framework for health status representation learning
  based on incorporating knowledge from similar patients.
\newblock In \emph{Proceedings of the AAAI conference on artificial
  intelligence}, volume~35, 715--723.

\bibitem[{Zhao et~al.(2023)Zhao, Yan, Sun, Xing, Meng, Wang, Cheng, Ren, and
  Yin}]{zhao2023knowing}
Zhao, Y.; Yan, L.; Sun, W.; Xing, G.; Meng, C.; Wang, S.; Cheng, Z.; Ren, Z.;
  and Yin, D. 2023.
\newblock Knowing What LLMs DO NOT Know: A Simple Yet Effective Self-Detection
  Method.
\newblock \emph{arXiv preprint arXiv:2310.17918}.

\bibitem[{Zhu et~al.(2024{\natexlab{a}})Zhu, Ren, Xie, Liu, Ji, Wang, Sun, He,
  Li, Zhu et~al.}]{zhu2024realm}
Zhu, Y.; Ren, C.; Xie, S.; Liu, S.; Ji, H.; Wang, Z.; Sun, T.; He, L.; Li, Z.;
  Zhu, X.; et~al. 2024{\natexlab{a}}.
\newblock REALM: RAG-Driven Enhancement of Multimodal Electronic Health Records
  Analysis via Large Language Models.
\newblock \emph{arXiv preprint arXiv:2402.07016}.

\bibitem[{Zhu et~al.(2024{\natexlab{b}})Zhu, Wang, Gao, Tong, An, Liao,
  Harrison, Ma, and Pan}]{zhu2024prompting}
Zhu, Y.; Wang, Z.; Gao, J.; Tong, Y.; An, J.; Liao, W.; Harrison, E.~M.; Ma,
  L.; and Pan, C. 2024{\natexlab{b}}.
\newblock Prompting Large Language Models for Zero-Shot Clinical Prediction
  with Structured Longitudinal Electronic Health Record Data.
\newblock \emph{arXiv preprint arXiv:2402.01713}.

\end{thebibliography}

\newpage
\section{Ethical Considerations and Limitations}
In this work, we focus on improving the interpretability of healthcare analysis by leveraging large language models (LLMs). We aim to provide a more comprehensive and accurate analysis of patients' conditions, which can help healthcare professionals make better decisions. However, there are several ethical considerations associated with our approach that warrant careful attention. 

First, the use of LLMs in healthcare analysis raises concerns about patient privacy and data security. We acknowledge that patient data is sensitive and must be handled carefully to ensure patient confidentiality. Therefore, we only apply open-source LLMs that can be inference locally. Using online services or cloud-based models may expose patient data to potential security risks.

Besides, the interpretability of LLMs is still an ongoing research area, and the generated analysis may not always be accurate or reliable. To this end, we generate multiple analyses and combine them with EHR modes rather than relying solely on LLMs. Moreover, the generated analyses may contain biases or errors, which could lead to incorrect conclusions. While IntelliCare aims to provide additional knowledge to existing models, it should not replace human judgment or clinical expertise. Physicians should utilize them as a reference rather than a definitive diagnosis or treatment plan. Patients should use IntelliCare as supplementary information and consult healthcare professionals for medical advice before making any decisions.

\section{Analysis of IntelliCare}
In this section, we provide an analysis of why IntelliCare can work effectively. For simplicity, we denote $X$ to represent the input variable of the EHR analysis task, which includes patient demographic information, conditions, surgical procedures, and medication regimens. Let $Y$ be the output variable, which can be the mortality rate, the probability of readmission, the length of stay, etc. Traditional EHR models aim to learn $P(Y|X)$ from the data, which is challenging due to the high-dimensional and small-sample characteristics of EHR data. In this work, we introduce additional knowledge variable $K$ from LLM to facilitate learning, that is to say, we learn the factorized form of $P(Y|X)$:
\begin{equation}\label{eq:p}
    P(Y|X) = P(Y|X,K)P(K|X),
\end{equation}
where the term $P(K|X)$ represents the process that the LLM outputs personalized knowledge given on $X$, while $P(Y|X, K)$ indicates the process that the predictor makes predictions by combining both the input $X$ and the knowledge $k$. If $P(K|X)$ is properly set up so that $K$ is highly relevant to both $X$ and $Y$, and can provide additional predictive power for $Y$, then learning $P(Y|X, K)$ will be simpler than learning $P(Y|X)$. However, directly letting the LLM generate $K$ based on $X$ might suffer from high variance due to the ambiguous problem, which leads to a high variance of $P(K|X)$. To mitigate this, we rely on the bias-variance tradeoff, introducing reasonable bias to reduce variance. Specifically, the bias we introduce is the assumption that the patient population has a clustered structure (such as in a Gaussian Mixture Model), which is a reasonable assumption in the real world. Under GMM, $P(X)$ can be expressed as:
\begin{equation}
    P(X) = \sum_{z=1}^{M} \pi_z \cdot \mathcal{N}(X | \mu_z, \Sigma_z),
\end{equation}
where $M$ is the number of mixture components, $\pi_z$ is the mixing coefficient for the $z$-th Gaussian distribution and $ \mathcal{N}(X | \mu_z, \Sigma_z) $ is the probability density function of the $z$-th Gaussian distribution with mean $ \mu_z $ and covariance matrix $\Sigma_z$. The posterior probability that input $x$ belongs to the $z$-th component can be given as:
\begin{equation}\label{eq:GMM}
    p(Z=z | X=x) = \frac{\pi_z \cdot \mathcal{N}(x | \mu_z, \Sigma_z)}{\sum_{j=1}^{M} \pi_j \cdot \mathcal{N}(x | \mu_j, \Sigma_j)}
\end{equation}
using the Bayes rule. Combining Eq \eqref{eq:GMM} with Eq \eqref{eq:p}, we get

\begin{equation}
\begin{aligned}
    P(Y|X) = \sum_{z=1}^{M} &P(Y|X,K) \cdot P(K|Z=z,X) \\
    & \cdot P(Z=z|X),
\end{aligned}
\end{equation}
where $P(Z=z|X)$ represents the process that we infer the cohort which $X$ belongs to, $P(K|Z=z, X)$ represents the process that the LLM outputs knowledge based on input and cohort information, which has lower variance due to the additional information.

Considering the inconsistency issue that LLM may output different knowledge for the same patient, we generate multiple knowledge $K=k$ for one patient $X$, and we get
\begin{equation}
    \begin{aligned}
        P(Y|X) = \sum_{k}\sum_{z=1}^{M} &P(Y|X,K=k) \cdot P(K=k|Z=z,X) \\
        & \cdot P(Z=z|X),
\end{aligned}
\end{equation}
where $P(K=k|Z=z, X)$ is the process that we infer multiple knowledge based on the cohort information and the input $X$. By introducing different knowledge, we can reduce the variance of the knowledge from the LLM and provide more reliable analysis for the patient.

In summary, we have transformed a difficult-to-learn $P(Y|X)$ into an easier-to-learn $P(Y|X, K)$ by incorporating the proper knowledge from LLMs. To reduce the variance in the knowledge output by the LLM, we additionally introduce a clustering bias while designing a cohort inference process to guide the output of the LLM using cohort information and generate multiple analyses for one patient.

\section{Implementation Details}
\subsection{Dataset Preprocessing}
When preprocessing datasets, we refer to the code in PyHealth \cite{yang2023pyhealth} and GraphCare \cite{jiang2024graphcare}. We convert the ICD-9-CM codes \footnote{\url{https://www.cdc.gov/nchs/icd/icd9cm.htm}} and ICD-10-CM codes \footnote{\url{https://www.cms.gov/medicare/icd-10/2023-icd-10-cm}} codes into CCS-CM codes \footnote{\url{https://www.accessdata.fda.gov/scripts/cder/ndc/index.cfm}} to capture condition concepts. Similarly, we map ICD-9-PROC codes and ICD-10-PCS codes \footnote{\url{https://www.cms.gov/medicare/icd-10/2023-icd-10-pcs}} to CCS-PROC \footnote{\url{https://www.cdc.gov/nchs/icd/icd9cm.htm}} codes for procedure concepts. For drug concepts, we convert
NDC \footnote{\url{https://www.accessdata.fda.gov/scripts/cder/ndc/index.cfm}} codes into level-3 ATC \footnote{\url{https://www.who.int/tools/atc-ddd-toolkit/atc-classification}} codes. This process speeds up training and improves performance by reducing the granularity of medical concepts.

\subsection{Knowledge Generation}

The following is a prompt template for generating analysis for a patient in the ICU. The template is designed to provide a comprehensive overview of the patient's condition, including the patient's demographics, diagnosis, treatment, and the statistics of the cohorts that the patient may belong to. We convert the medical codes to their corresponding concepts to make them can be understood by LLMs.

\renewcommand{\dblfloatpagefraction}{.9}
\begin{table*}
    \caption{Performance and standard deviations on the MIMIC-III dataset using Mistral-7B-Instruct.}
    \label{result_mistral}
    \centering
    \small
    \setlength{\tabcolsep}{4pt}
    \begin{tabular}{lcccccccc}
        \toprule
        \multirow{3}{*}{\textbf{Model}} & \multicolumn{2}{c}{\textbf{Mortality Prediction}} & \multicolumn{2}{c}{\textbf{Readmission Prediction}} & \multicolumn{4}{c}{\textbf{Length-of-Stay Prediction}}\\
        \cmidrule(lr){2-3} \cmidrule(lr){4-5} \cmidrule(lr){6-9}
        & AUROC(\%) & AUPRC(\%) & AUROC(\%) & AUPRC(\%) & AUROC(\%) & Kappa(\%) & Accuracy(\%) & F1 Score(\%)\\
        \midrule
        StageNet+IntelliCare & 64.68{\scriptsize $\pm$0.69} & 18.19{\scriptsize $\pm$0.28} & 60.46{\scriptsize $\pm$1.18} & 38.71{\scriptsize $\pm$0.53} & 78.31{\scriptsize $\pm$0.18} & 32.83{\scriptsize $\pm$0.99} & 38.95{\scriptsize $\pm$0.71} & 25.67{\scriptsize $\pm$0.85} \\
        ConCare+IntelliCare & 65.95{\scriptsize $\pm$1.01} & 18.41{\scriptsize $\pm$1.07} & 61.67{\scriptsize $\pm$1.38} & 39.62{\scriptsize $\pm$1.18} & 80.33{\scriptsize $\pm$0.37} & 35.53{\scriptsize $\pm$0.65} & 41.21{\scriptsize $\pm$0.93} & 28.08{\scriptsize $\pm$0.80} \\
        GRASP+IntelliCare & 64.13{\scriptsize $\pm$1.55} & 16.76{\scriptsize $\pm$0.69} & 60.55{\scriptsize $\pm$1.60} & 37.56{\scriptsize $\pm$0.82} & 79.59{\scriptsize $\pm$0.61} & 34.75{\scriptsize $\pm$0.93} & 41.02{\scriptsize $\pm$0.80} & 28.11{\scriptsize $\pm$1.31} \\
        SAFARI+IntelliCare & 67.79{\scriptsize $\pm$0.80} & 19.07{\scriptsize $\pm$0.87} & 60.01{\scriptsize $\pm$0.40} & 37.53{\scriptsize $\pm$0.40} & 79.21{\scriptsize $\pm$0.52} & 35.03{\scriptsize $\pm$0.72} & 40.99{\scriptsize $\pm$0.65} & 27.16{\scriptsize $\pm$0.96} \\
        \bottomrule
    \end{tabular}
\end{table*}

\begin{table*}
    \caption{Performance and standard deviations on the MIMIC-IV dataset using Mistral-7B-Instruct.}
    \label{result_mistral1}
    \centering
    \small
    \setlength{\tabcolsep}{4pt}
    \begin{tabular}{lcccccccc}
        \toprule
        \multirow{3}{*}{\textbf{Model}} & \multicolumn{2}{c}{\textbf{Mortality Prediction}} & \multicolumn{2}{c}{\textbf{Readmission Prediction}} & \multicolumn{4}{c}{\textbf{Length-of-Stay Prediction}}\\
        \cmidrule(lr){2-3} \cmidrule(lr){4-5} \cmidrule(lr){6-9}
        & AUROC(\%) & AUPRC(\%) & AUROC(\%) & AUPRC(\%) & AUROC(\%) & Kappa(\%) & Accuracy(\%) & F1 Score(\%)\\
        \midrule
        StageNet+IntelliCare & 74.30{\scriptsize $\pm$1.04} & 7.94{\scriptsize $\pm$1.37} & 67.63{\scriptsize $\pm$0.28} & 57.50{\scriptsize $\pm$0.49} & 73.69{\scriptsize $\pm$0.29} & 25.42{\scriptsize $\pm$0.95} & 29.41{\scriptsize $\pm$1.01} & 18.68{\scriptsize $\pm$0.94} \\
        ConCare+IntelliCare & 76.71{\scriptsize $\pm$0.73} & 8.81{\scriptsize $\pm$0.42} & 68.38{\scriptsize $\pm$0.28} & 58.57{\scriptsize $\pm$0.43} & 74.67{\scriptsize $\pm$0.19} & 25.20{\scriptsize $\pm$0.37} & 30.61{\scriptsize $\pm$0.51} & 19.91{\scriptsize $\pm$0.35} \\
        GRASP+IntelliCare & 72.78{\scriptsize $\pm$2.16} & 7.74{\scriptsize $\pm$1.49} & 68.18{\scriptsize $\pm$0.27} & 58.43{\scriptsize $\pm$0.59} & 73.97{\scriptsize $\pm$0.10} & 25.53{\scriptsize $\pm$0.03} & 29.46{\scriptsize $\pm$0.29} & 18.94{\scriptsize $\pm$0.21} \\
        SAFARI+IntelliCare & 78.11{\scriptsize $\pm$1.38} & 10.73{\scriptsize $\pm$1.22} & 67.87{\scriptsize $\pm$0.56} & 57.05{\scriptsize $\pm$0.99} & 72.54{\scriptsize $\pm$0.54} & 23.70{\scriptsize $\pm$0.67} & 28.22{\scriptsize $\pm$0.85} & 17.19{\scriptsize $\pm$0.84} \\
        \bottomrule
    \end{tabular}
\end{table*}

\begin{figure*}[!h]
    \centering
    \includegraphics[width=2\columnwidth]{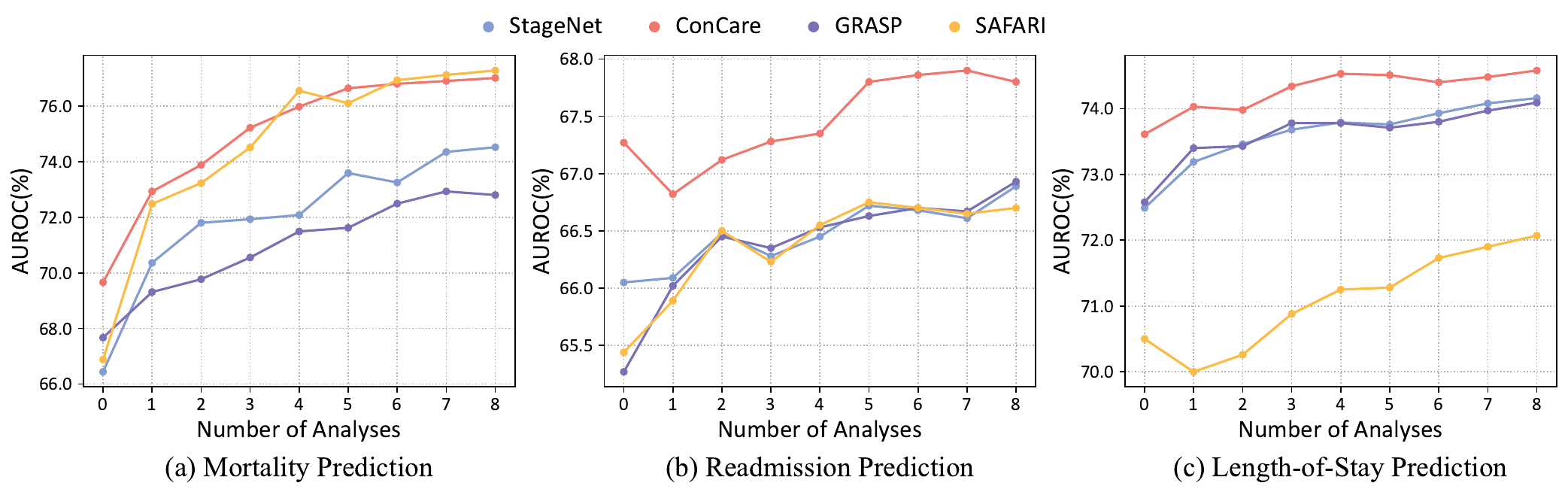}
    \caption{Ablation comparison of IntelliCare for different numbers of analyses on the MIMIC-IV dataset.}
    \label{k-mimiciv}
\end{figure*}

\begin{tcolorbox}[colback=gray!10,
    colframe=black,
    width=\columnwidth,
    arc=2mm, auto outer arc,
    title={Prompt of Generating Analysis},breakable,]
    There is a \textit{age}-year-old \textit{male/female} patient who is admitted to the ICU.

    This patient is diagnosed with \textit{condition\_num} \textbf{conditions}:
    \dots

    The patient has undergone \textit{procedure\_num} \textbf{procedures}:
    \dots

    This patient used \textit{medication\_num} \textbf{medications}:
    \dots

    The statistics of the cohorts that the patient may belong to (for reference only) are:

    - Cohort 1 (\textit{prob}\% probability): The average age is \textit{cohort\_age} years old, \textit{cohort\_gender}\% of the patients are male, and the average number of conditions, procedures, and medications are \textit{cohort\_condition\_num}, \textit{cohort\_procedure\_num}, and \textit{cohort\_medication\_num}, respectively. The average length of stay is \textit{cohort\_los} days. The readmission rate in 30 days is \textit{cohort\_readmission}. The mortality rate at the next admission is \textit{cohort\_mortality}.

    - \dots

    Based on the patient information provided above, please briefly summarize the patient's condition. Assess risks with brief explanations for length of stay, 30-day readmission, and outcome of next admission. 
    Your analysis should reflect the unique aspects of this patient's condition and treatment.
\end{tcolorbox}

When generating the patient-level knowledge, we set the maximum generation length to 1024, which can meet the vast majority of generation lengths. Since the context length of LLaMA-3 is 8192, we drop the drug information if the length of the prompt exceeds.

\subsection{Model Training}
To fairly compare different approaches, we set the embedding dimension for the medical codes to 32, and the hidden dimension to 128 since these settings have been shown to work well in previous studies and in our experiments.
For StageNet, ConCare, GRASP, and SAFARI, we train them with a learning rate of 1e-3 and a batch size of 256. We train the models for 20 epochs, and the best model is selected based on the validation set. Since GraphCare needs huge GPU RAM capacity and training time (about 4 hours for one runtime), we train them for 15 epochs, with a learning rate of 2.5e-4. When training IntelliCare, we freeze the parameters of the EHR models and use a learning rate of 1e-4 for 20 epochs with a batch size of 256 for StageNet, ConCare, GRASP, and SAFARI. When using GraphCare, we train IntelliCare for 15 epochs with a learning rate of 1.25e-5 and batch size of 16.
Adam optimizer \cite{kingma2014adam} is used for all models. We use the same hyperparameters for IntelliCare on both the MIMIC-III and MIMIC-IV datasets.

All the experiment was done in a machine equipped with CPU: Intel Xeon CPU E5-2650 v4, 256GB RAM, and GPU: NVIDIA GeForce RTX 3090, 
using Pytorch 2.3.0.

\section{Additional Experimental Results}
\subsection{Results with Mistral-7B-Instruct}
We provide the results via Mistral-7B-Instruct in Table \ref{result_mistral} and Table \ref{result_mistral1}. We can see that IntelliCare improves the performance of existing models on both the MIMIC-III and MIMIC-IV datasets. The results are consistent with the findings using LLaMA-3 8B Instruct. By comparing the results between the two LLMs, we find that Mistral-7B-Instruct achieves better performance on the length-of-stay prediction task. This shows that IntelliCare is robust and can be applied to different LLMs to improve healthcare analysis.

\subsection{Results with Different Numbers of Analyses on the MIMIC-IV Dataset}
We illustrate the ablation comparison of IntelliCare for different numbers of analyses on the MIMIC-IV dataset in Figure \ref{k-mimiciv}. We can see that as the number of analyses increases, the performance of IntelliCare improves on all tasks. This finding is consistent with the results on the MIMIC-III dataset. The results demonstrate that IntelliCare can provide more accurate and reliable analysis by generating multiple analyses for each patient.

\section{Case Study of the Effects of Different Prompts}
In this section, we provide cases to show the difference between prompts. Given a de-identified patient, its demographics, conditions, procedures, medications, outcomes, and the statistics of the cohorts are as follows:
\begin{tcolorbox}[colback=gray!10,
    colframe=black,
    width=\columnwidth,
    arc=2mm, auto outer arc,
    title={Patient Medical Records},breakable,]
    There is a 44-year-old male patient who is admitted to the ICU since gastrointestinal bleed.

    This patient is diagnosed with 9 conditions:
    
    - Acute vascular insufficiency of intestine
    
    - Anoxic brain damage
    
    \dots

    The patient has undergone 7 procedures:

    - Open and other right hemicolectomy

    - Continuous invasive mechanical ventilation for less than 96 consecutive hours

    \dots

    This patient used 47 medications:

    - Famotidine (IV)
    
    - Chlorpromazine HCl

    \dots

    The patient's \textbf{length of stay is 15 days}. This patient \textbf{readmitted within 30 days} and \textbf{discharged alive at the next admission}.

    The statistics of the cohorts that the patient may belong to (for reference only) are:

    - Cohort 1 (71\% probability): The average age is 67 years old, 67\% of the patients are male, and the average number of diagnoses, procedures, and medications are 12, 5, and 35, respectively. The average length of stay is 12 days. The readmission rate in 30 days is 57\%. The mortality rate at the next admission is 16\%.
    
    - Cohort 2 (17\% probability): The average age is 80 years old, 55\% of the patients are male, and the average number of diagnoses, procedures, and medications are 16, 9, and 54, respectively. The average length of stay is 24 days. The readmission rate in 30 days is 24\%. The mortality rate at the next admission is 21\%.
    
    - Cohort 3 (11\% probability): The average age is 64 years old, 54\% of the patients are male, and the average number of diagnoses, procedures, and medications are 17, 13, and 67, respectively. The average length of stay is 40 days. The readmission rate in 30 days is 31\%. The mortality rate at the next admission is 18\%.

\end{tcolorbox}

\subsection{Analysis Case without Task-Relevant Instructions}
Without task-relevant instructions, the generated analysis with the highest average relevance to the EHR encoder (ConCare) is as follows:

\begin{tcolorbox}[colback=gray!10,
    colframe=black,
    width=\columnwidth,
    arc=2mm, auto outer arc,
    title={Analysis from IntelliCare$_{t-}$},breakable,]
Based on the patient information provided, the 44-year-old male patient is admitted to the ICU with a diagnosis of gastrointestinal bleed, which is likely caused by the presence of Acute vascular insufficiency of intestine and Other suppurative peritonitis. He also has several other comorbidities, including Congestive heart failure, Anoxic brain damage, Pneumonia, and Chronic respiratory failure.

The patient has undergone multiple procedures, including a right hemicolectomy, endoscopy of the small intestine, and placement of a venous catheter, indicating the presence of a complex and potentially severe illness.

The patient is prescribed 47 medications, which is an unusually high number. The medications include a wide range of drugs to manage the patient's symptoms, including acid reflux, nausea, pain, and seizures. Some medications, such as Epoetin Alfa and Gentamicin Sulfate, suggest the presence of an underlying infection or bone marrow suppression.

The patient's condition is characterized by a high level of complexity, with multiple comorbidities and medications. The patient's ICU admission suggests that he is at high risk of developing complications, such as sepsis, respiratory failure, and cardiac arrest. The patient's medical team will need to carefully monitor his condition and adjust his treatment plan accordingly to prevent further deterioration.

\end{tcolorbox}

We can see that the generated analysis without task-relevant instructions provides a general overview of the patient's condition but lacks specific details and insights. The analysis does not provide a detailed assessment of the patient's risks or outcomes, which limits its usefulness for clinical prediction and decision-making.

\subsection{Analysis Case with Task-Relevant instructions but without Cohort Statistics}
Without the cohort statistics, the generated analysis with the highest average relevance to the EHR encoder (ConCare) is as follows:

\begin{tcolorbox}[colback=gray!10,
    colframe=black,
    width=\columnwidth,
    arc=2mm, auto outer arc,
    title={Analysis from IntelliCare$_{c-}$},breakable,]
Based on the patient's condition, it appears that he has a complex and critical illness with multiple comorbidities and concurrent medical conditions. The patient has a recent history of gastrointestinal bleed, and their conditions, including acute vascular insufficiency of the intestine, pneumonia, congestive heart failure, and anoxic brain damage, indicate a high risk of morbidity and mortality.

The patient has undergone multiple procedures, including open and other right hemicolectomy, continuous invasive mechanical ventilation, and parenteral infusion of concentrated nutritional substances, which highlights the need for close monitoring and management.


Assessing risks for length of stay, 30-day readmission, and outcome of next admission:

\textbf{Length of stay}: The patient's complex condition, multiple comorbidities, and need for continued ventilation and nutritional support suggest a \textbf{prolonged hospital stay}.

\textbf{30-day readmission}: Given the patient's underlying conditions and the need for ongoing management, there is a \textbf{high risk of readmission within 30 days}. The patient may require ongoing monitoring and treatment for their chronic conditions, such as congestive heart failure and chronic respiratory failure.

\textbf{Outcome of next admission}: Given the patient's critical condition, \textbf{there is a risk of prolonged hospitalization, readmission, and even mortality}. However, with aggressive management and close monitoring, the patient may stabilize and recover, potentially requiring continued treatment and management.
\end{tcolorbox}

With task-relevant instructions, we can find that the LLM provides a comprehensive assessment of the patient's condition, risks, and outcomes. However, LLM usually only provides fuzzy evaluation, which may be inaccurate or unreliable. This fuzzy assessment has a large variance, which can create greater uncertainty for subsequent fusion.

\subsection{Analysis Case with Cohort Statistics}
With the cohort statistics, the generated analysis with the highest average relevance to the EHR encoder (ConCare) is as follows:

\begin{tcolorbox}[colback=gray!10,
    colframe=black,
    width=\columnwidth,
    arc=2mm, auto outer arc,
    title={Analysis from IntelliCare},breakable,]
    Based on the patient's medical record, the patient is a 44-year-old male admitted to the ICU with gastrointestinal bleed and diagnosed with 9 conditions, including acute vascular insufficiency of the intestine, congestive heart failure, anoxic brain damage, and chronic respiratory failure. He underwent 7 procedures, including open and other right hemicolectomy, and received 47 medications.

    Assessing risks, I would predict that the patient's \textbf{length of stay will be longer than average, likely exceeding 12 days}, due to the severity of his conditions and the need for prolonged mechanical ventilation. \textbf{His 30-day readmission risk is also higher than average, likely exceeding 57\%}, given the complexity of his medical condition and the likelihood of ongoing complications. Regarding the outcome of the next admission, \textbf{the patient's mortality rate is difficult to predict} without more specific information about his underlying conditions and the effectiveness of his treatment. 

    In summary, the patient is a high-risk patient with a complex medical condition, requiring prolonged ICU care and multiple procedures. His 30-day readmission risk is high, and his mortality rate at the next admission is uncertain. Close monitoring and management of his condition, as well as careful consideration of medication and procedure risks, will be essential to optimize his outcomes.

\end{tcolorbox}
 
By comparing these analyses, we find that the generated analysis provides a more detailed assessment of the patient's condition, risks, and outcomes when incorporating the cohort statistics. The analysis considers the patient's unique characteristics and compares them to the average statistics of the cohorts, providing a more personalized and accurate evaluation. Since the analysis involves quantitative assessment, it is more reliable and has less variance.

\section{Case of LLM's Knowledge at the Medical Code Level}
We provide a case to demonstrate LLMs knowledge at the medical code level, both in the form of knowledge graph (triples) and natural language. 

\begin{tcolorbox}[colback=gray!10,
    colframe=black,
    width=\columnwidth,
    arc=2mm, auto outer arc,
    title={Knowledge in the Form of Knowledge Graph},breakable,]



    (Acetaminophen, treats, Pain)

    (Acetaminophen, treats, Fever)


    (Acetaminophen, has\_side\_effect, Nausea)

    (Acetaminophen, has\_side\_effect, Rash)










    (Acetaminophen, potential\_risk, Liver\_damage)

    (Acetaminophen, interacts\_with, Warfarin)

    (Acetaminophen, has\_pharmacological\_class, Aniline\_analgesics)

    (Acetaminophen, used\_in\_treatment\_of, Osteoarthritis)
\end{tcolorbox}

\begin{tcolorbox}[colback=gray!10,
    colframe=black,
    width=\columnwidth,
    arc=2mm, auto outer arc,
    title={Knowledge in the Form of Natural Language},breakable,]
    Acetaminophen (also known as paracetamol outside North America) is a widely used over-the-counter analgesic (pain reliever) and antipyretic (fever reducer). It is often used to treat headaches, muscle aches, arthritis, backaches, toothaches, colds, and fevers.

    Acetaminophen works by inhibiting the synthesis of prostaglandins in the central nervous system and through a peripheral action by blocking pain-impulse generation. It acts centrally on the hypothalamic heat-regulating center to produce antipyresis. Unlike nonsteroidal anti-inflammatory drugs (NSAIDs), it has minimal anti-inflammatory properties.
    
    Therapeutic Uses

    - Pain Relief: Effective for mild to moderate pain such as headaches, menstrual periods, toothaches, backaches, and osteoarthritis.
    
    - Fever Reduction: Commonly used to reduce fever in conditions like colds and flu.
    
    Dosage and Administration
    
    - Adults: Typically, the dosage ranges from 325 mg to 650 mg every 4 to 6 hours, not exceeding 4,000 mg per day.

    - Children: Dosage is based on the child's weight and age, with pediatric formulations available.

    Acetaminophen is generally well-tolerated when used as directed. Common side effects may include: nausea, rash, headache, liver toxicity.

    Overdose of acetaminophen can cause severe liver damage. Symptoms of overdose include: nausea, vomiting, sweating, general malaise, and elevated liver enzymes. In severe cases, overdose can lead to liver failure and death. Immediate medical attention is required, and N-acetylcysteine (NAC) is the antidote used to treat acetaminophen toxicity.
\end{tcolorbox}

Knowledge graphs and natural language descriptions provide structured information about acetaminophen, including its therapeutic uses, side effects, etc. However, for specific patients, some of this knowledge might be irrelevant, requiring additional knowledge filtering methods to be more effective. Furthermore, knowledge generated by LLMs may not include less frequently used information, even though it might be crucial for certain patients.

\end{document}